\documentclass[10pt,twocolumn,letterpaper]{article}

\usepackage{cvpr}              

\definecolor{cvprblue}{rgb}{0.21,0.49,0.74}
\usepackage[pagebackref,breaklinks,colorlinks,allcolors=cvprblue]{hyperref}

\usepackage{graphicx}    
\usepackage{subcaption}  
\usepackage{booktabs} 
\usepackage{multirow}
\usepackage{array} 
\usepackage{kotex}

\usepackage{xcolor}

\usepackage[capitalize]{cleveref}
\crefname{section}{Sec.}{Secs.}
\Crefname{section}{Section}{Sections}
\Crefname{table}{Table}{Tables}
\crefname{table}{Tab.}{Tabs.}

\def\paperID{8247} 
\def\confName{CVPR}
\def\confYear{2025}

\title{Devil is in the Detail: Towards Injecting Fine Details of Image Prompt in Image Generation via Conflict-free Guidance and Stratified Attention}

\author{Kyungmin Jo \hspace{2.7cm}
Jooyeol Yun \hspace{2.7cm}
Jaegul Choo \\
Korea Advanced Institute of Science and Technology (KAIST) \\
Daejeon, Korea\\
{\tt\small \{bttkm, blizzard072, jchoo\}@kaist.ac.kr} 
}

\begin{document}

\twocolumn[{%
\renewcommand\twocolumn[1][]{#1}%
\maketitle
\begin{center}
    \centering
    \captionsetup{type=figure}
    \includegraphics[width=\textwidth]{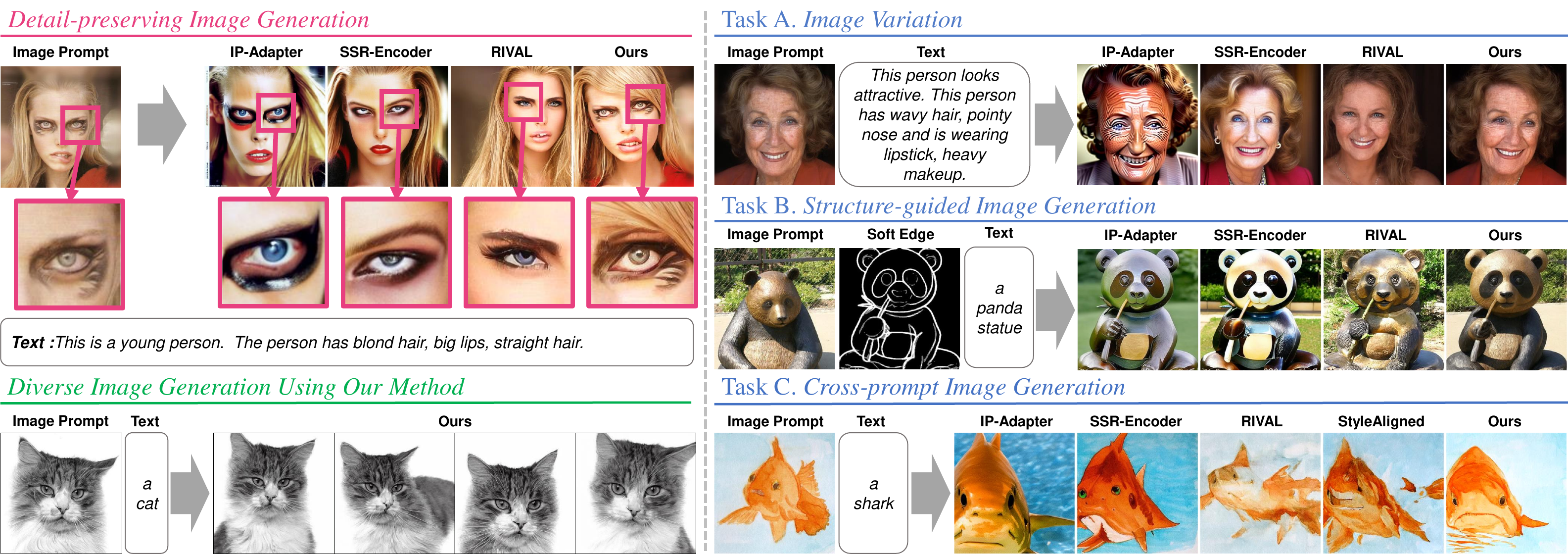}
    \caption{
    Our method faithfully \textit{reflects the details} of image prompts (\eg, eye makeup) for \textit{diverse} image generation across \textit{various tasks} (Tasks A-C) in a training-free manner. In contrast, existing methods often fail to incorporate fine details of the image prompt. Even with training, IP-Adapter~\cite{ye2023ip} and SSR-Encoder~\cite{zhang2024ssr} struggle to properly represent the color and texture of the image prompt. RIVAL~\cite{zhang2024real} and StyleAligned~\cite{hertz2024style} also fail to generate high-fidelity images that faithfully reflect the image or text prompts due to a trade-off between the realism of the generated results and alignment with the image prompt. Extra results are in the Supplementary (Suppl.).
    }
    \label{fig_teaser}
\end{center}%
}]

\maketitle
\begin{abstract}

While large-scale text-to-image diffusion models enable the generation of high-quality, diverse images from text prompts, these prompts struggle to capture intricate details, such as textures, preventing the user intent from being reflected. This limitation has led to efforts to generate images conditioned on user-provided images, referred to as image prompts. Recent work modifies the self-attention mechanism to impose image conditions in generated images by replacing or concatenating the keys and values from the image prompt. This enables the self-attention layer to work like a cross-attention layer, generally used to incorporate text prompts.
In this paper, we identify two common issues in existing methods of modifying self-attention that hinder diffusion models from reflecting the image prompt. By addressing these issues, we propose a novel method that generates images that properly reflect the details of image prompts. First, existing approaches often neglect the importance of image prompts in classifier-free guidance, which directs the model towards the intended conditions and away from those undesirable. Specifically, current methods use image prompts as both desired and undesired conditions, causing conflicting signals. To resolve this, we propose conflict-free guidance by using image prompts only as desired conditions, ensuring that the generated image faithfully reflects the image prompt.
In addition, we observe that the two most common self-attention modifications involve a trade-off between the realism of the generated image and alignment with the image prompt, achieved by selectively using keys and values from both images. Specifically, selecting more keys and values from the image prompt improves alignment, while selecting more from the generated image enhances realism. To balance both, we propose an alternative self-attention modification method, Stratified Attention, which jointly uses keys and values from both images rather than selecting between them.
Through extensive experiments across three distinct image generation tasks, we demonstrate that the proposed method outperforms existing image-prompting models in faithfully reflecting the image prompt.
Our code is available at \href{https://github.com/bttkm82/InDetail-IP}{https://github.com/bttkm82/InDetail-IP}.

\end{abstract}    
\section{Introduction}
\label{sec_intro}

The emergence of large-scale text-to-image diffusion models~\cite{rombach2022high, podell2023sdxl} has enabled users to generate diverse images from text. However, describing complex scenes using only text prompts can be complicated, leading to ongoing efforts to incorporate user-provided image conditions to better reflect user intent~\cite{ye2023ip,zhang2024ssr,wang2024instantid,ruiz2023dreambooth}. As part of these efforts, modifying the self-attention mechanism in pre-trained text-to-image diffusion models has recently attracted interest in areas such as image editing~\cite{tumanyan2023plug,cao2023masactrl,liu2024towards}, story visualization~\cite{zhou2024storydiffusion}, image variation~\cite{zhang2024real}, and style transfer~\cite{chung2024style}, as a flexible method for incorporating user-provided image conditions into generated images. 
This modification is typically achieved by integrating the keys and values of the condition in the self-attention layer, as the keys and values derived from the image features contain rich details about images.

Since the keys and values embody the context of images, these techniques are particularly powerful for conditioning diffusion models on \emph{image prompts}, which contain the intricate details of a scene (\eg, the fine patterns of eye makeup in \cref{fig_teaser}) that go beyond the level of details text descriptions can reach. 
Specifically, recent findings~\cite{chung2024style, hertz2024style, zhang2024real} show that replacing or concatenating the new keys and values (KV) of the generated images to those of the image prompt in the self-attention layer simulates a cross-attention with the image prompt, functioning similarly to the way text prompts affect image generation.
This simulated cross-attention offers key benefits over training an additional cross-attention layer~\cite{ye2023ip}: it eliminates the need for retraining with each new diffusion model and minimizes information loss from using smaller datasets for the training, as self-attention features are generated within the same model.

\begin{figure}[t!]
    \centering
    \includegraphics[width=0.9\linewidth]{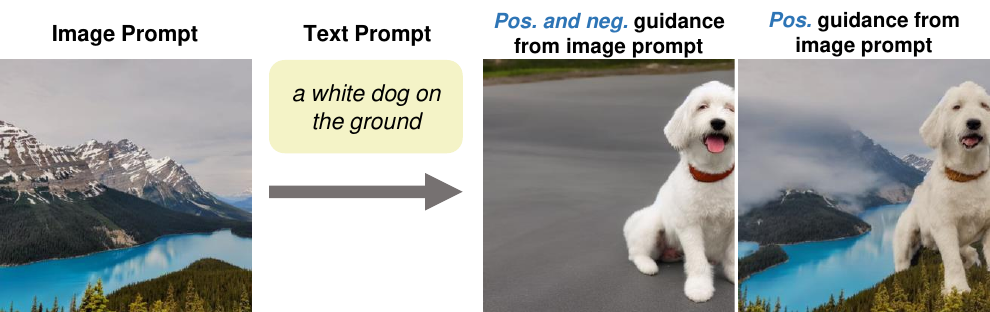} 
    \caption{Using the image prompt as \textit{positive} guidance ensures that it is reflected in the generated images, but using it as \textit{both} positive (pos.) and negative (neg.) guidance fails due to the conflicting guidance.}
    \vspace{-5mm}
    \label{fig_rmneg_img}
\end{figure}

However, we find that existing self-attention alterations-based approaches overlook an important aspect, which is the role of image prompts during \emph{classifier-free guidance}~\cite{ho2021classifier}. 
Classifier-free guidance steers the diffusion model toward desired conditions (\ie, positive guidance) and away from undesired ones (\ie, negative guidance), enhancing the alignment between generated images and the user-provided conditions.
Current approaches use image prompts to estimate both positive-guidance and negative-guidance scores, causing image prompts to influence the negative scores as well. This sends a strong signal to the model to \textit{avoid} reflecting the image prompt, thereby compromising alignment between the generated image and the image prompt.
The effect of this conflicting guidance is illustrated in~\cref{fig_rmneg_img}, where the details of the image prompt are often lost or even ignored.
Based on this insight, we propose conflict-free guidance by adjusting classifier-free guidance to \emph{exclude} image prompts from estimating negative-guidance scores, enabling generations that faithfully reflect the provided image prompt.

Furthermore, we observe that two common self-attention modifications, which we call KV replacement and KV concatenation, represent a trade-off between the realism of the generated results and alignment with the image prompt. KV replacement ensures strong alignment with the image prompt but struggles to generate realistic attributes not specified in the image prompt (\eg, frontal view of a cat), as it relies solely on the keys and values from the image prompt (\eg, side view of a cat) while discarding those from the generated results. In contrast, KV concatenation, which combines keys and values from both the image prompt and the generated results, enhances realism but reduces the alignment due to attention bias arising from the queries being derived from the same features as the generated keys and values. 
To achieve both realism in the generated results and alignment with the image prompt, we propose Stratified Attention (\cref{fig_attn_4types_stratattn}) as a new self-attention modification that effectively addresses this bias in KV concatenation.
Specifically, we calculate the attention scores for image prompts and generated results separately, rather than computing them all at once. The scores calculated in this way are then aggregated afterward.
This facilitates the image prompts to attend to relevant features while preserving the original attention flow of the generated image. 

To sum up, our key contributions are as follows:
\begin{itemize}
    \item We correct the classifier-free guidance in modified self-attention by using the image prompt exclusively as positive guidance, rather than as conflicting guidance, to faithfully reflect the image prompt in image generation.
    \item We propose Stratified Attention as a new self-attention modification to effectively reflect the image prompt by ensuring a balanced attention flow between the image prompt and the generated image.
    \item Through extensive experiments, 
    our method achieves state-of-the-art performance in image prompting, both qualitatively and quantitatively.
\end{itemize}

\begin{figure}[t!]
    \centering
    \includegraphics[width=\linewidth]{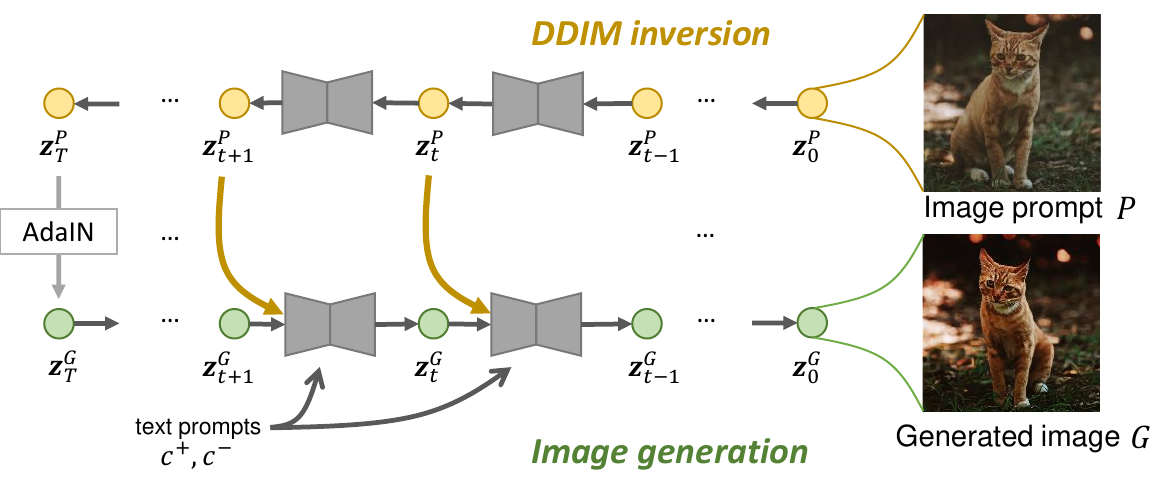}
    \caption{Framework of the self-attention modification for image prompt: Image $G$ is generated from the initial noised latent $\textbf{z}_T^G$ using the initial inverted noised latent $\textbf{z}_T^P$, which is obtained from the image prompt $R$ via the DDIM inversion. 
    }
    \vspace{-3mm}
    \label{fig_framework}
\end{figure}

\section{Related Work}
\label{sec_related_work}

\paragraph{Self-attention Modification in diffusion models.}
In diffusion models, KV replacement and KV concatenation are commonly used self-attention modifications.
KV replacement (\cref{fig_attn_4types_replace}) is primarily employed in tasks that require strong incorporation of the given image, such as image editing~\cite{cao2023masactrl} and style transfer~\cite{chung2024style}. Conversely, KV concatenation (\cref{fig_attn_4types_concat})
is utilized to tasks that require diverse image generation while reflecting the given image, such as style-aligned image generation~\cite{hertz2024style}, image variation~\cite{zhang2024real}, and story visualization~\cite{zhou2024storydiffusion}.
However, KV concatenation weakens alignment with the given image due to attention bias from queries sharing features with generated keys and values. To address this, StyleAligned~\cite{hertz2024style} weights the image prompt’s attention scores, while RIVAL~\cite{zhang2024real} and StoryDiffusion~\cite{zhou2024storydiffusion} use KV replacement at specific time steps. Although these methods improve alignment, they sacrifice image realism by limiting the incorporation of generated features.
To solve this problem, our proposed Stratified Attention separately computes attention scores for the image prompt and generated results, thereby enhancing image prompt incorporation without sacrificing generated feature integration.

\paragraph{Image-prompting diffusion models.}
In text-to-image diffusion models, there are two distinct mechanisms to condition the generation process on an ``image prompt'': 1) cross-attention and 2) self-attention.
Cross-attention-based approaches~\cite{xu2024prompt,ye2023ip,zhang2024ssr} integrate image prompts using auxiliary encoders or cross-attention layers, requiring additional training. However, the smaller datasets used for this training, compared to those used for diffusion models, can limit the quality of the generated images (\eg, the image variation results of IP-Adapter in~\cref{fig_teaser}).
In contrast, self-attention-based approaches\cite{zhang2024real} incorporate the image prompt through the aforementioned self-attention alteration without training, allowing them to fully utilize the image generation capabilities of diffusion models. 
However, while cross-attention-based approaches direct image prompts solely to the positive branch during classifier-free guidance, similar to text prompts, self-attention-based methods apply the same image prompt to both branches, disregarding classifier-free guidance principles. This improper use of classifier-free guidance prevents the details of the image prompt from being reflected in the generated images. In our method, we correct the classifier-free guidance in self-attention modification, enhancing the incorporation of the image prompt.  

\begin{figure}[t!]
    \centering
    \begin{subfigure}[b]{0.47\linewidth}    
        \centering
        \includegraphics[width=\linewidth]{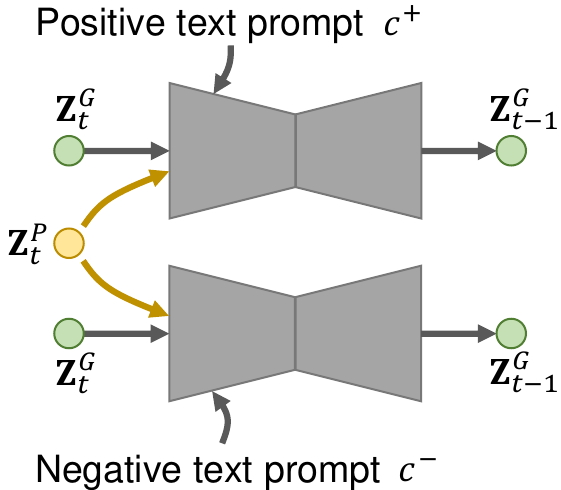}
        \caption{Conflicting guidance}
        \label{fig_rm_negprompt_prev}
    \end{subfigure}
    \hfill
    \begin{subfigure}[b]{0.47\linewidth}    
        \centering
        \includegraphics[width=\linewidth]{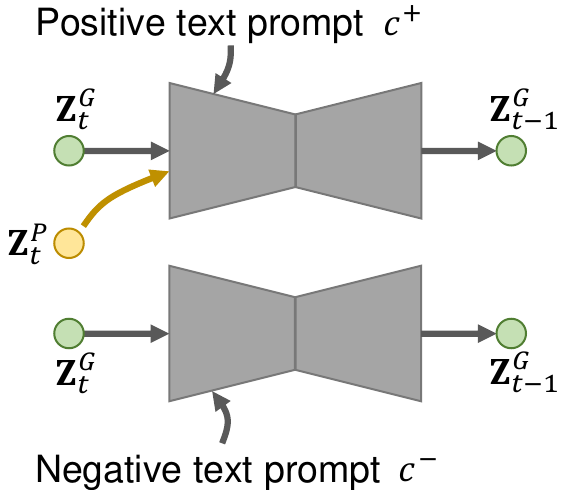}
        \caption{Conflict-free guidance (Ours)}
        \label{fig_rm_negprompt_propose}
    \end{subfigure}
    \caption{
    Illustration of the classifier-free guidance when image prompts are (a) used and (b) omitted during the estimation of the negative guidance scores. 
    }
    \vspace{-5mm}
    \label{fig_rm_negprompt}
\end{figure}

\section{Preliminaries}
\label{sec_preliminaries}
\paragraph{Attention Fusion}
Existing studies~\cite{cao2023masactrl, zhang2024real} that incorporate a given image into the generated output via controlling key and values in self-attention layer selectively apply different types of self-attention (\cref{fig_attn_4types}) across time steps, employing a technique called the Attention Fusion strategy. These studies apply original self-attention or KV concatenation along with KV replacement to generate realistic images by utilizing the features of the generated image, rather than relying solely on the features of the given image, while still reflecting the given image.
For image prompting, RIVAL~\cite{zhang2024real} combines KV replacement and KV concatenation, referred to as "cross-image self-attention injection" in their paper. In this process, the attention process of KV concatenation is as follows:
\begin{equation}
    \label{eq_attn_concat}
    \begin{aligned}
    & \text{Attn}(M_t,V_t)=M_t\cdot (V_t^G \oplus V_t^P), \\
    & \text{where}\ M_t=\text{softmax}(\frac{Q_t^G ({K_t^G \oplus K_t^P})^{T}}{ \sqrt{d}}),
    \end{aligned}
\end{equation}
$M_t$ and $V_t$ are the attention map and values, respectively, at time step $t$. $\oplus$ indicates concatenation along the spatial dimension, and $d$ is the latent projection dimension of the key $K$ and query $Q$.

\paragraph{Self-attention-based Image Prompting.}
As illustrated in~\cref{fig_framework}, generating images that reflect the image prompt $P$ through the self-attention modification is structured into two distinct stages: DDIM inversion~\cite{song2020denoising} and image generation.
In the DDIM inversion stage, we obtain the inverted noised latent chain $\{\textbf{z}_T^P, \cdots, \textbf{z}_0^P\}$ from the image prompt $P$ through a deterministic DDIM inversion. The inverted noised latent $\textbf{z}_t^P$ from this chain is provided to the model at the corresponding time step $t$ during image generation, playing a crucial role in incorporating the features of the image prompt $P$ into the generated images $G$ through the self-attention alterations. 

In the image generation stage, we create images $G$ from randomly sampled Gaussian noise $\textbf{z}$, conditioned on the positive text prompt $c^+$, the negative text prompt $c^-$, and the inverted noised latent $\textbf{z}_T^P$. 
To transfer the color tone of the image prompt $P$ to the generated image $G$, we apply AdaIN to the Gaussian noise $\textbf{z}$ using the initial inverted noised latent $\textbf{z}_T^P$, as utilized in StyleID~\cite{chung2024style}, to derive the initial noised latent $\textbf{z}_T^G$. This process is defined as follows.
\begin{equation}
\begin{aligned}
\textbf{z}_T^{G} &= \text{AdaIN}(\textbf{z}, \textbf{z}_T^P), \\
\text{where } \text{AdaIN}(x, y) &= \sigma(y) \left(\frac{x - \mu(x)}{\sigma(x)} \right) + \mu(y).
\label{eq_init_latent}
\end{aligned}
\end{equation}
$\mu(\cdot)$ and $\sigma(\cdot)$ represent the mean and standard deviation computed for each channel, respectively.
To utilize classifier-free guidance~\cite{ho2021classifier}, we duplicate the noised latent $\textbf{z}_t^G$ at each time step and provide each copy to the model along with the positive text prompt $c^+$ and the negative text prompt $c^-$, respectively. At this stage, to incorporate the image prompt $P$ into the generated image $G$, the model also supplies the previously obtained inverted noised latent $\textbf{z}_t^P$ along with the noised latent $\textbf{z}_t^G$. 
\begin{figure}[t!]
    \centering
    \includegraphics[width=\linewidth]{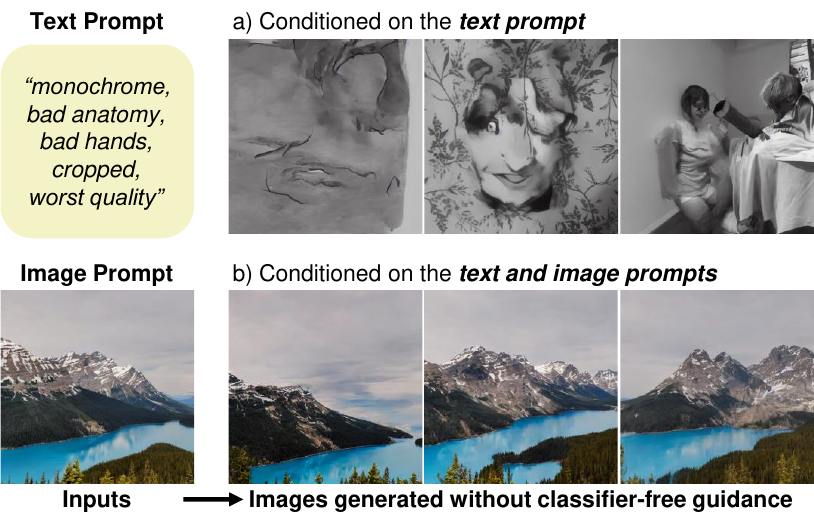}
    \caption{
    To compare the impact of negative guidance between a) text-to-image diffusion models and b) self-attention modification-based image-prompting models, we generate images using negative prompts commonly applied in each model, without employing classifier-free guidance. Unlike negative guidance from text prompts, negative guidance from image prompts through self-attention modification incorporates the image prompt, thereby guiding the model to move away from the image prompt.
    }
    \vspace{-5mm}
    \label{fig_analysis_sacontrol}
\end{figure}

\section{Method}
\label{sec_method}
We introduce a novel method that generates images faithfully reflecting given prompts by controlling keys and values in the self-attention layers of text-to-image diffusion models without training. This section details two key components to enhance image prompt alignment: 1) conflict-free guidance (\cref{sec_rm_negprompt}), and 2) Stratified Attention (\cref{sec_stratattn}).

\begin{figure}[t!]
    \centering
    \includegraphics[width=\linewidth]{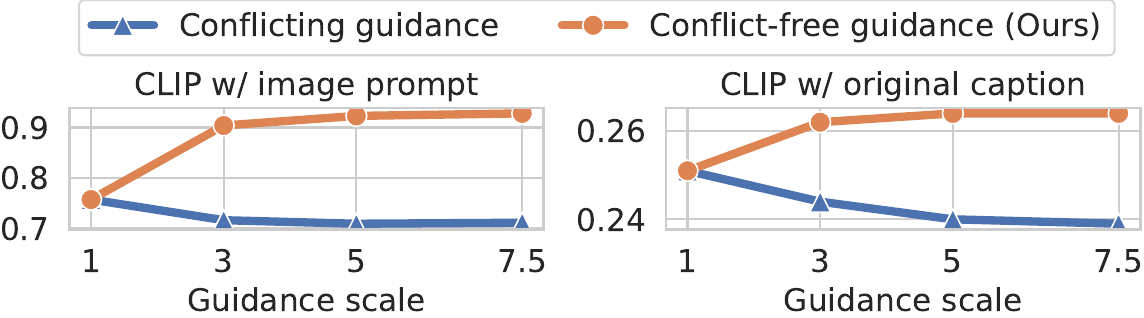}
    \caption{Employing the image prompt as both positive and negative guidance fails to effectively direct the model to reflect the image prompt due to conflicting guidance (\textit{blue line}), whereas using it exclusively as positive guidance provides clearer direction by eliminating this conflict (\textit{orange line}).
    }
    \vspace{-5mm}
    \label{fig_rmneg_graph}
\end{figure}

\subsection{Conflict-free Guidance}
\label{sec_rm_negprompt}

As shown in~\cref{fig_rmneg_img}, the self-attention modification used in existing image-prompting research~\cite{zhang2024real} hinders the incorporation of the image prompt in the generated images due to conflicting guidance, where the image prompt is applied as both positive and negative guidance (\cref{fig_rm_negprompt_prev}) during classifier-free guidance. To resolve this conflict, we correct the guidance term for image prompts by using them exclusively as positive guidance (\cref{fig_rm_negprompt_propose}), based on two analyses.

To investigate the impact of using an image prompt as negative guidance within the self-attention modification on the generated images, we first compare images produced from prompts typically employed for negative guidance in both the original diffusion model and self-attention modification-based image-prompting methods. In this experiment, to assess the influence of prompts used for negative guidance, we generate images without applying classifier-free guidance, using negative prompts in place of positive prompts. 
As shown in \cref{fig_analysis_sacontrol}, unlike images generated from negative text prompts, the negative guidance in self-attention modification models incorporates the image prompt, which guides the model away from reflecting the image prompt.

\begin{figure*}[ht!]
    \centering
    \begin{subfigure}[b]{0.208\textwidth}    
        \centering
        \includegraphics[width=\textwidth]{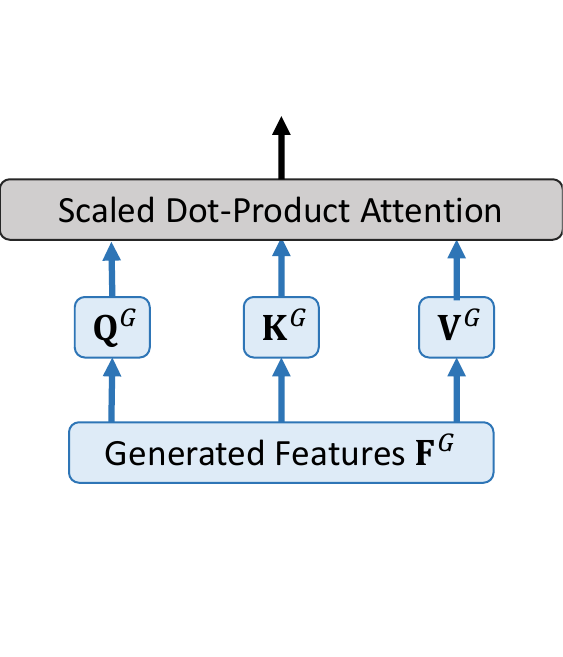}
        \caption{Original self-attention}
        \label{fig_attn_4types_org}
    \end{subfigure}
    \hfill
    \begin{subfigure}[b]{0.208\textwidth}    
        \centering
        \includegraphics[width=\textwidth]{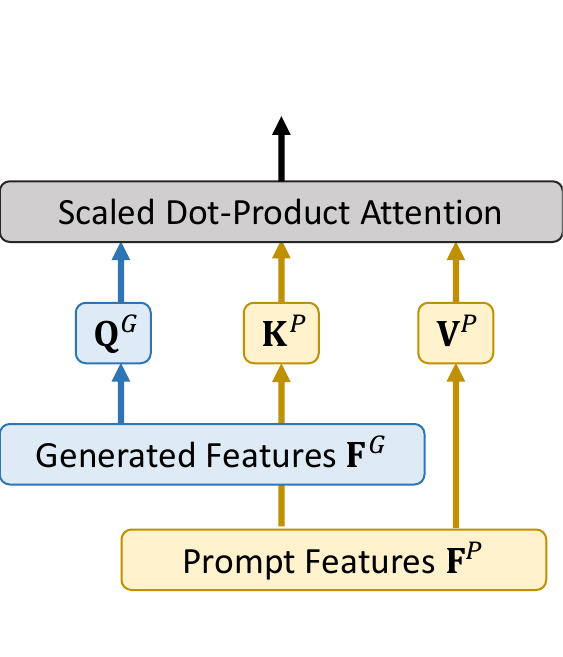}
        \caption{KV replacement}
        \label{fig_attn_4types_replace}
    \end{subfigure}
    \hfill
    \begin{subfigure}[b]{0.208\textwidth}    
        \centering
        \includegraphics[width=\textwidth]{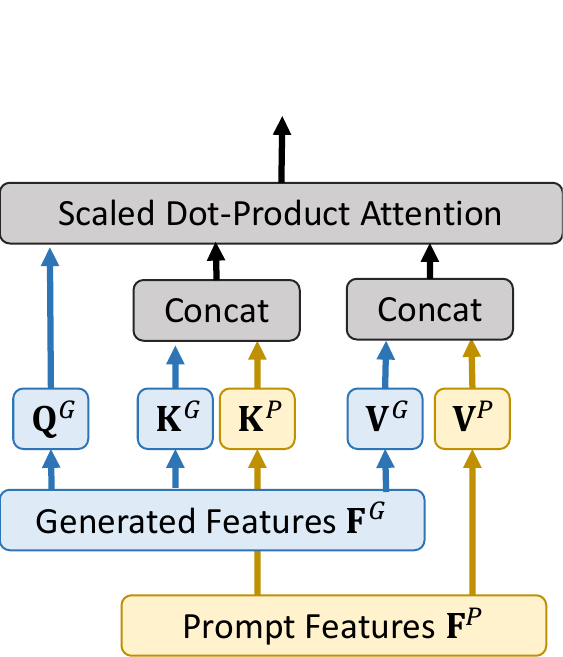}
        \caption{KV concatenation}
        \label{fig_attn_4types_concat}
    \end{subfigure}
    \hfill
    \begin{subfigure}[b]{0.312\textwidth}
        \centering
        \includegraphics[width=\textwidth]{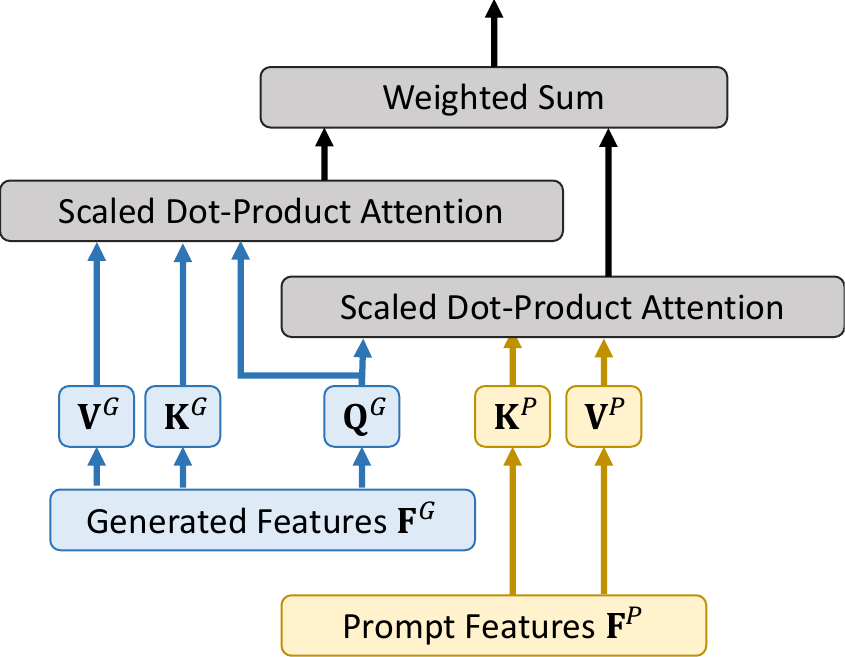}
        \caption{Stratified attention}
        \label{fig_attn_4types_stratattn}
    \end{subfigure}
    
    \caption{Types of attention modification techniques. (a) Original self-attention (b) KV replacement: The keys and values from the image prompts replace the original keys and values. (c) KV concatenation: The key and values from the image prompts are concatenated with the original key and values. (d) Stratified attention: The attention scores are calculated independently and summed afterward.}
    \label{fig_attn_4types}
\end{figure*}

In addition, during classifier-free guidance, we examine the effect of excluding the image prompt as negative guidance in the self-attention modification on incorporation of the image prompt into generated images. Specifically, we incrementally increase the guidance scale and calculate CLIP scores~\cite{radford2021learning} to evaluate the alignment of the generated images with the image prompt and the ground truth text. To isolate the effect of the image prompt, a null prompt is used as the positive text prompt during image generation. As shown in \cref{fig_rmneg_graph}, excluding the image prompt from negative guidance leads to a more faithful incorporation of it than including it as negative guidance. This demonstrates that providing conflict-free guidance by using the image prompt solely as a positive prompt in self-attention modification plays an important role in ensuring the generated image faithfully reflects the image prompt.

\subsection{Stratified Attention}
\label{sec_stratattn}
To generate diverse and realistic images while reflecting the image prompt, RIVAL~\cite{zhang2024real} employs Attention Fusion, selectively applying KV replacement (\cref{fig_attn_4types_replace}) and KV concatenation (\cref{fig_attn_4types_concat}) rather than relying solely on KV replacement.
While this strategy enables the incorporation of both the generated image and the image prompt features via KV concatenation, we observe that KV concatenation tends to favor higher attention scores for the generated features over those of the image prompt (\cref{fig_scorediff}), which reduces alignment with the image prompt.

To overcome these limitations, we propose Stratified Attention (\cref{fig_attn_4types_stratattn}). While KV concatenation computes the attention process for both generated features and image prompt features in an integrated manner (\cref{eq_attn_concat}), Stratified Attention separately calculates the attention processes for each feature and then combines them through a weighted sum as follows:
\begin{equation}
    \label{eq_attn_stratattn}
    \begin{aligned}
        \text{Attn}(M_t^G, M_t^P, V_t^G, V_t^P) = & \lambda_G M_t^G V_t^G + \lambda_P M_t^P V_t^P, \\
        \text{where }M_t= & \text{softmax} \left( \frac{Q_t ({K_t})^{\intercal}}{\sqrt{d}} \right).
    \end{aligned}
\end{equation}
This separation of operations ensures a balanced attention flow between both sources. 
Instead of KV replacement, we use Stratified Attention with KV concatenation in Attention Fusion to consistently reflect both generated and image prompt features across all time steps, ensuring a balanced reflection of each.

\begin{figure}[t!]
    \centering
    \includegraphics[width=\linewidth]{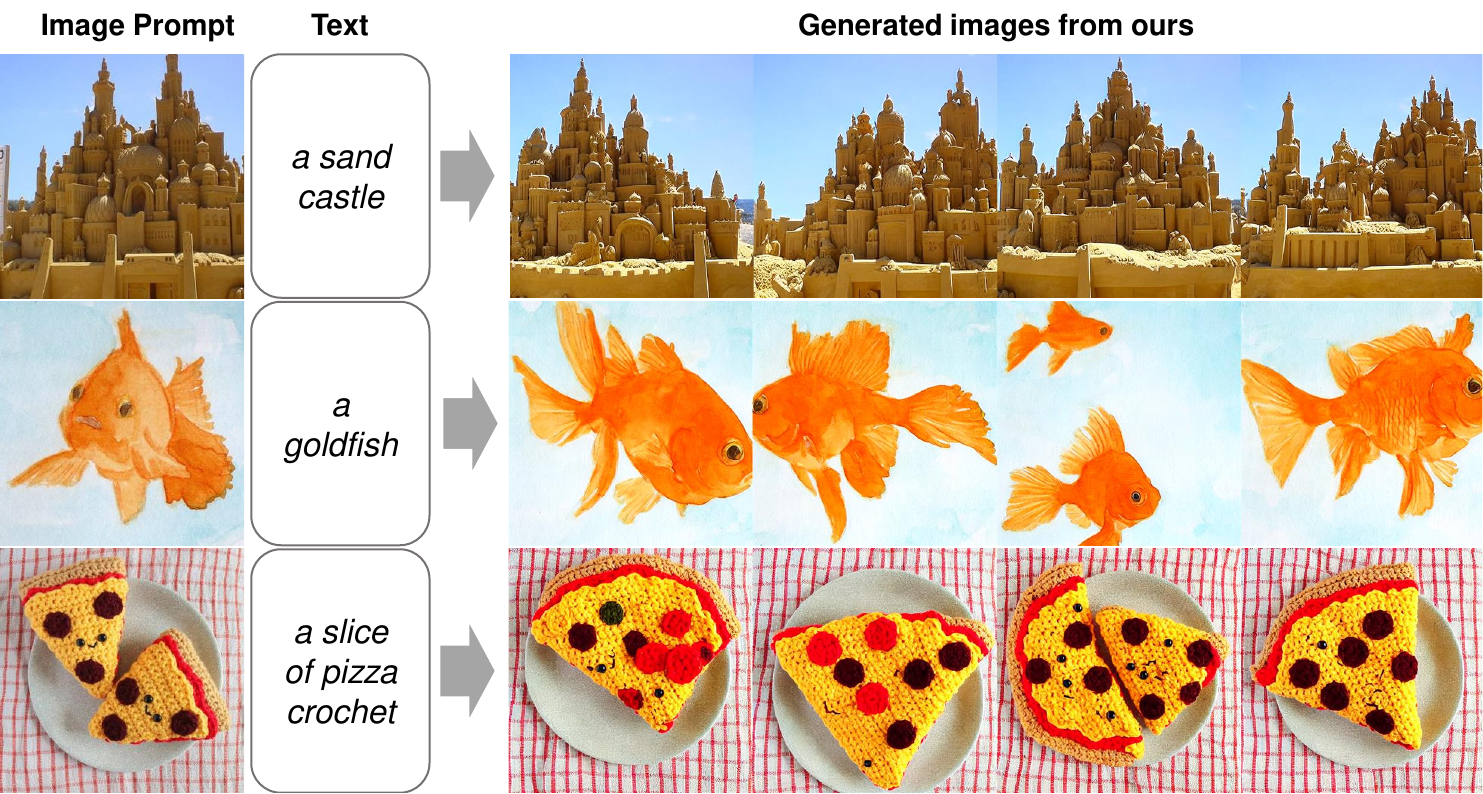}
    \caption{Our method generates images that not only reflect the image prompt but also exhibit diverse structures.}
    \vspace{-5mm}
    \label{fig_imgvar_ours}
\end{figure}

\begin{figure*}[ht]
    \centering
    \begin{subfigure}{0.43\textwidth}
        \includegraphics[width=\linewidth]{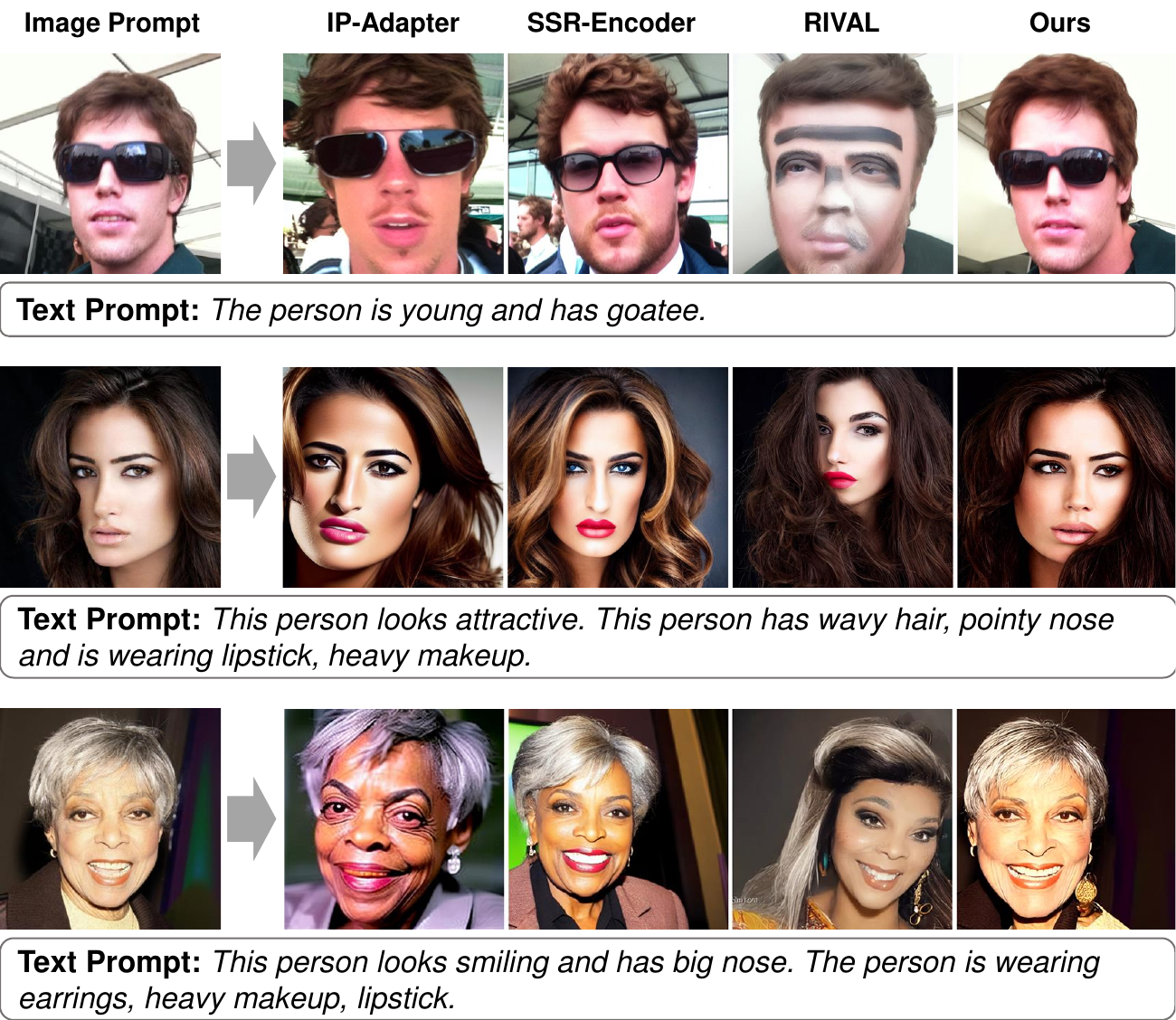}
        \caption{
        Image variation. Unlike ours, baselines often fail to reflect details of image prompts (\eg, color of lipstick, type of sunglasses).
        }
        \label{fig_comp_imgvar}
    \end{subfigure}
    \hfill
    \begin{subfigure}{0.55\textwidth}
        \includegraphics[width=\linewidth]{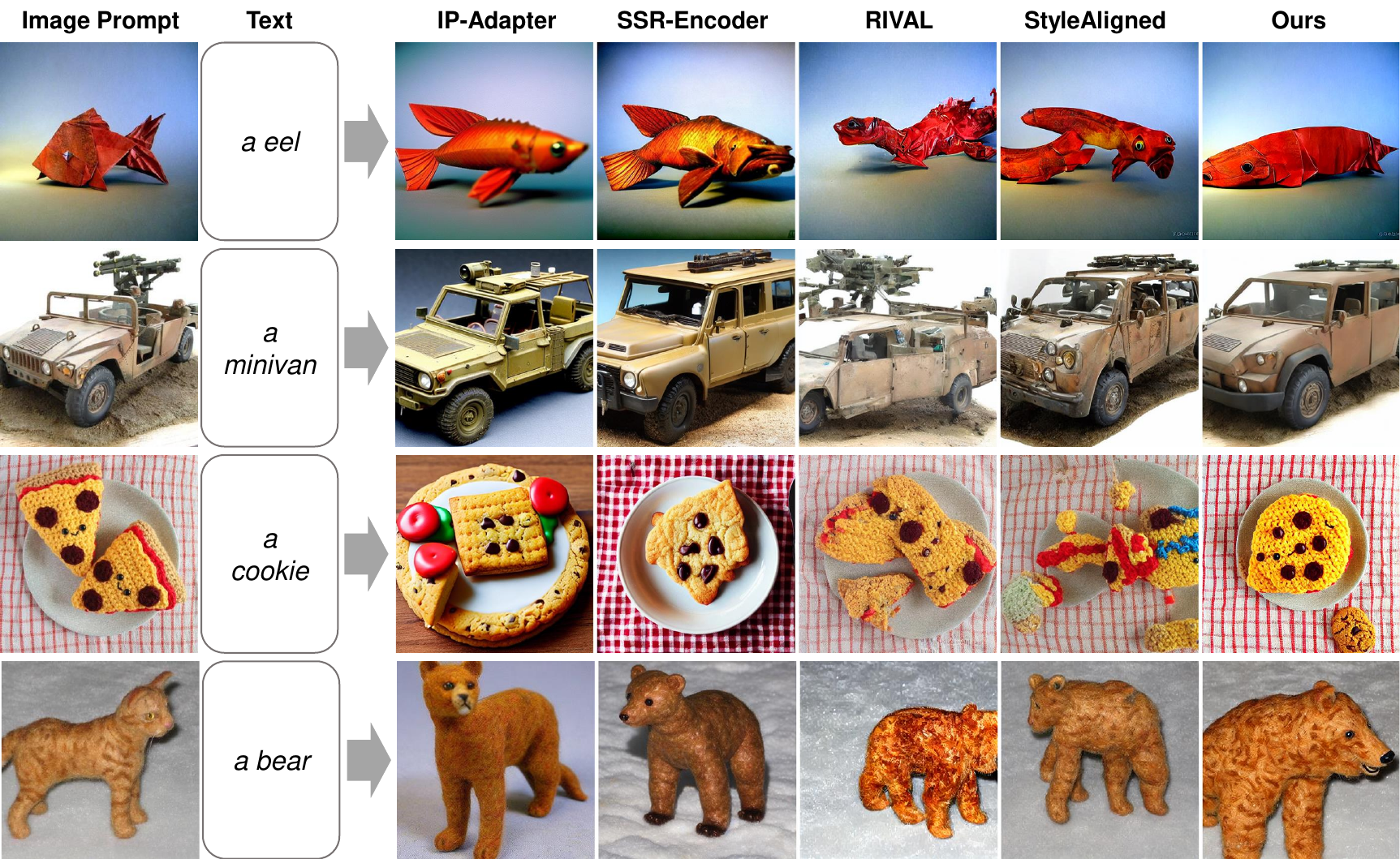}
        \caption{
        Cross-prompt image generation. In general, models generate images that reflect the content of the text prompt and the style or texture of the image prompt. While baseline models often fail to accurately reflect either the image or the text prompt, our method successfully integrates both prompts in a harmonious way.
        }
        \label{fig_comp_style}
    \end{subfigure}
    \vskip\baselineskip
        \centering
    \begin{subfigure}{0.49\textwidth}
        \includegraphics[width=\linewidth]{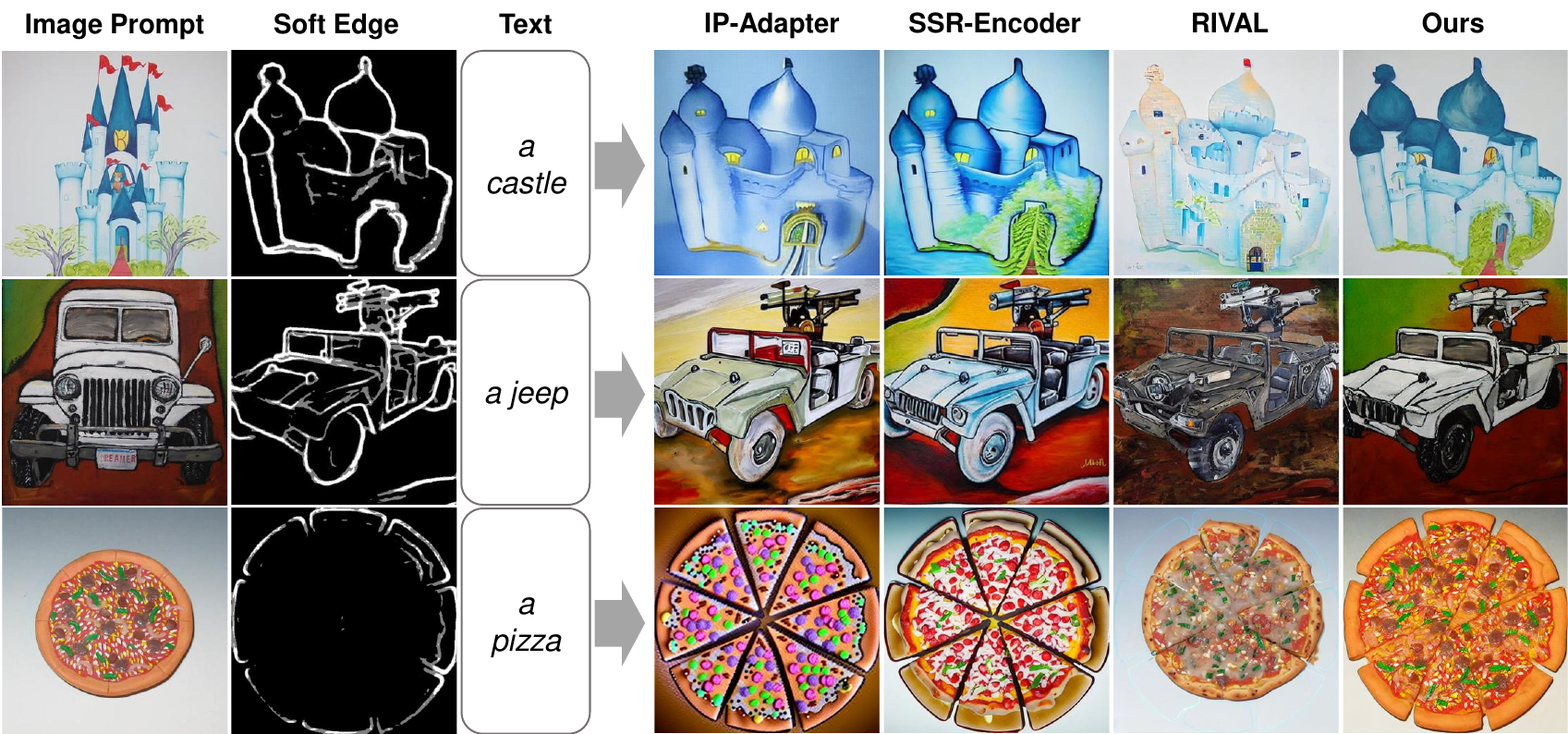}
        \caption{
        Structure-guided image generation using soft edges. Baseline methods exhibit color shifts or fail to incorporate the structure of soft edges.
        }
        \label{fig_comp_cn_depth}
    \end{subfigure}
    \hfill
    \begin{subfigure}{0.49\textwidth}
        \includegraphics[width=\linewidth]{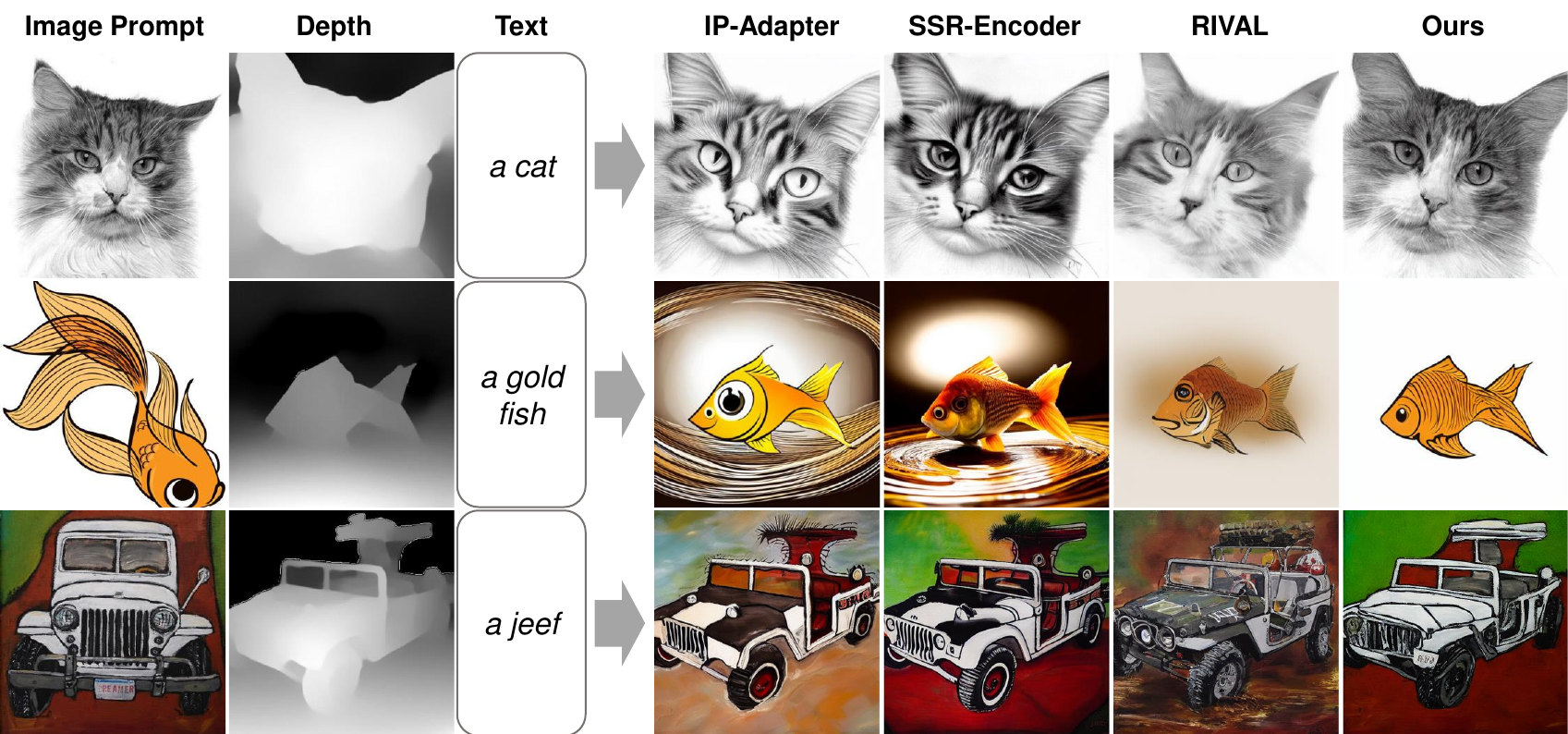}
        \caption{
        Structure-guided image generation using depth. Baseline methods often fail to capture the color tones and textures of the image prompt.
        }
        \label{fig_comp_cn_softedge}
    \end{subfigure}
    \caption{Qualitative comparison with baselines across three tasks: (a) Image variation, (b) Cross-prompt image generation, and (c-d) Structure-guided image generation. Additional results are in the Supplementary (Suppl.)
    }
    \vspace{-5mm}
    \label{fig:four_images}
\end{figure*}

\section{Experiments}
\label{sec_exp_results}

\paragraph{Evaluation Metrics}
We quantitatively compare our model to image-prompting approaches across four aspects.
For image prompting, we measure the image-alignment by calculating the similarity using image encoders such as CLIP (CLIP-I)~\cite{radford2021learning}.
Additionally, similar to text-to-image generation, we measure text-alignment and the realism using CLIP scores (CLIP-T) and Fr\'echet Inception Distance (FID)~\cite{heusel2017gans}. 
However, since both CLIP scores can remain high even when the generated images closely replicate the image prompt, we also assess structural diversity between the generated images by evaluating structural distance based on the DINO self-similarity score (DINO-G)~\cite{caron2021emerging}.
To evaluate the metrics, we generate four images for each image prompt.

\paragraph{Baselines}
For our comparisons, we utilize three image-prompting models as baselines: IP-Adapter~\cite{ye2023ip}, SSR-Encoder~\cite{zhang2024ssr} and RIVAL~\cite{zhang2024real}. IP-Adapter and SSR-Encoder attach separate trainable adapters to the diffusion model, enabling it to reflect the image prompt in the generated images. 
In contrast, RIVAL is a training-free model that generates images aligned with the provided image prompt by the self-attention modification, which is frequently employed to reflect the given image. 
To ensure a fair comparison, we employ Stable Diffusion~\cite{rombach2022high} version 1.5 for all models including ours. Also, we use the same Gaussian noise, except for RIVAL, which shuffles the inverted noised latent $\textbf{z}_T^P$ to produce $\textbf{z}_T^G$.

\paragraph{Datasets}
We validate our approach on various text-image pairs from three datasets, COCO 2017~\cite{lin2014microsoft}, ImageNet-R-TI2I~\cite{tumanyan2023plug}, and CelebAHQ~\cite{karras2017progressive}.
Specifically, we utilize the validation set of COCO and randomly sample 100 images from the CelebAHQ dataset. For text prompts, we use the original captions from the COCO and the CelebAHQ dataset, while the text prompts for the ImageNet-R-TI2I dataset are generated using BLIP~\cite{li2022blip}.

\subsection{Comparison with Baselines}
\label{sec_comp_baseline}
In this section, we demonstrate the effectiveness of our proposed method by conducting qualitative and quantitative comparisons of its performance against baseline models across three tasks: 1) image variation, 2) cross-prompt image generation, and 3) structure-guided image generation. Extra qualitative results, implementation details, and ablation studies are in Suppl..

\begin{table*}[t]
    \centering
    \resizebox{\textwidth}{!}{%
    \begin{tabular}{@{}l|ccc|ccc|cccc@{}}
    \toprule
     & \multicolumn{3}{c|}{CelebAHQ} & \multicolumn{3}{c|}{ImageNet-R-TI2I} & \multicolumn{4}{c}{COCO validation} \\ \midrule
     & CLIP-I $\uparrow$ & CLIP-T $\uparrow$ & DINO-G $\uparrow$ & CLIP-I $\uparrow$ & CLIP-T $\uparrow$ & DINO-G $\uparrow$ & CLIP-I $\uparrow$ & CLIP-T $\uparrow$ & DINO-G $\uparrow$ & FID $\downarrow$ \\ \midrule
    IP-Adapter~\cite{ye2023ip} & 0.845 & 0.264 & 0.073 & \underline{0.870} & 0.300 & 0.092 & \underline{0.858} & 0.323 & \underline{0.089} & 15.485 \\
    SSR-Encoder~\cite{zhang2024ssr} & \underline{0.850} & 0.269 & 0.078 & 0.861 & 0.306 & 0.100 & 0.854 & 0.321 & 0.087 & 15.730 \\
    RIVAL~\cite{zhang2024real} & 0.739 & \textbf{0.289} & \textbf{0.095} & 0.836 & \textbf{0.317} & \underline{0.140} & 0.832 & \underline{0.325} & \underline{0.089} & \textbf{14.016} \\
    Ours & \textbf{0.878} & \underline{0.274} & \underline{0.094} & \textbf{0.883} & \underline{0.311} & \textbf{0.142} & \textbf{0.873} & \textbf{0.326} & \textbf{0.093} & \underline{14.557} \\ \bottomrule
    \end{tabular}%
    }
    \caption{Quantitative comparison with baseline models in image variation tasks. Our method shows competitive performance compared to baselines in structural diversity, image realism, and reflecting text prompts, while surpassing them in image prompt alignment, as measured by CLIP-I. Best and second best results are in \textbf{bold} and \underline{underlined}.}
    \vspace{-2mm}
    \label{table_imgvar}
\end{table*}
\begin{table*}[t]
\centering
\begin{minipage}{0.4\textwidth}
    \centering
    \includegraphics[width=\textwidth]{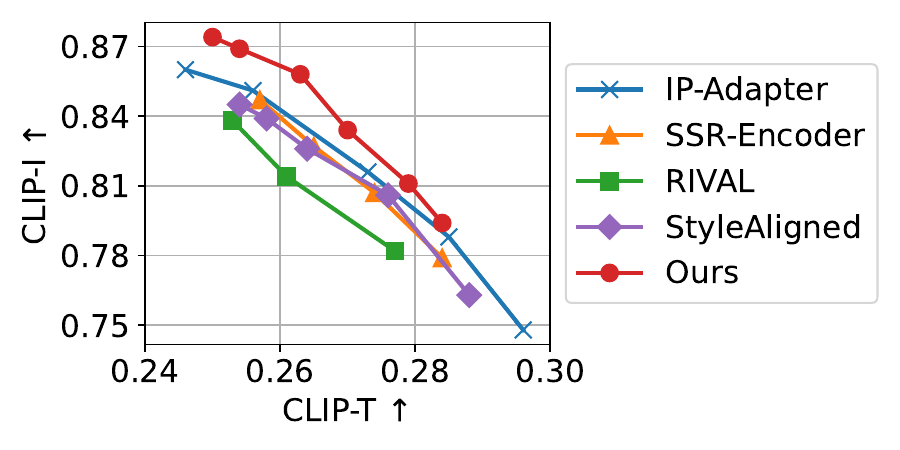}
    \vspace{-8mm}
    \captionof{figure}{Quantitative results for cross-prompt image generation. Each dot represents a different factor for adjusting the image prompt alignment. Our approach demonstrates high similarity in both image and text.}
    \vspace{-5mm}
    \label{fig_saig_graph}
\end{minipage}%
\hfill
\centering
\begin{minipage}{0.58\textwidth} 
    \centering
    \resizebox{\linewidth}{!}{%
    \begin{tabular}{@{}c|ccc|ccc@{}}
    \toprule
     & \multicolumn{3}{c|}{Depth} & \multicolumn{3}{c}{Soft Edge} \\ \midrule
     & CLIP-I $\uparrow$ & CLIP-T $\uparrow$ & DINO-C $\downarrow$ & CLIP-I $\uparrow$ & CLIP-T $\uparrow$ & DINO-C $\downarrow$\\ \midrule
    IP-Adapter & \underline{0.826} & 0.306 & \textbf{0.053} & \underline{0.811} & \underline{0.300} & \textbf{0.044} \\
    SSR-Encoder & 0.816 & \underline{0.309} & \underline{0.057} & 0.799 & 0.299 & \underline{0.052} \\
    RIVAL & 0.794 & 0.217 & 0.079 & 0.793 & 0.216 & 0.060 \\
    Ours & \textbf{0.861} & \textbf{0.310} & 0.066 & \textbf{0.844} & \textbf{0.308} & \underline{0.052} \\ \bottomrule
    \end{tabular}%
    }
    \caption{Quantitative comparison of structure-guided image generation. Our method demonstrates competitive performance with baselines in reflecting structural conditions based on DINO-C, while excelling in the integration of both image and text prompts based on CLIP scores. Best and second best results are in \textbf{bold} and \underline{underlined}.}
    \vspace{-5mm}
    \label{table_controlnet}
\end{minipage}
\end{table*}

\paragraph{Image Variation}
\label{sec_exp_imgvar}
We evaluate image variation performance across three datasets using four metrics: CLIP-I, CLIP-T, DINO-G, and FID. Due to the limited number of images in the ImageNet-R-TI2I and CelebA-HQ datasets, we compute FID only for the COCO validation set. The qualitative results in \cref{fig_comp_imgvar} show that baseline models generate images that capture the semantic meaning of the image prompt (\eg, woman) but fail to incorporate finer details (\eg, makeup color tone or type of sunglasses). In contrast, our method generates images that faithfully reflect these details. 

Along with these results, we provide qualitative examples that highlight our model's ability to generate diverse images from a single prompt. As shown in \cref{fig_imgvar_ours}, our approach consistently produces images with varied structures that remain well-aligned with the image prompt, ensuring coherent content across all outputs.
In addition, we quantitatively compare our method with the baselines in \cref{table_imgvar}. Our method performs competitively with the baselines in terms of text alignment, structural diversity, and image realism, while providing the most faithful alignment with the image prompt across multiple datasets.

\paragraph{Cross-prompt Image Generation}
\label{sec_exp_style}
Since image-prompting models are fundamentally based on text-to-image diffusion models, several methods~\cite{ye2023ip, zhang2024real, zhang2024ssr} provide functionalities to adjust the balance between text and image prompts during generation. Thus, we evaluate each model's ability to harmoniously integrate both prompts in the generated images by adjusting their hyperparameter settings and comparing the results. For this analysis, we set the text and image prompts with differing content to examine which of the two prompts (text or image) is reflected in the generated images. Given the difficulty of text prompts in defining styles such as texture, we set a content disparity between the text and image prompts for this experiment by using image-text editing pairs from the ImageNet-R-TI2I dataset. Also, we include StyleAligned~\cite{hertz2024style} as an additional baseline, which generates images from cross-prompt inputs. For quantitatively comparison, we evaluate performance in terms of image and text alignment based on two CLIP scores while varying the hyperparameters. 

As shown in~\cref{fig_saig_graph}, our method achieves higher alignment with both prompts compared to the baselines. Also, we visually display the results that show the best performance for each model (the closest result to the top-right corner of the graph in~\cref{fig_saig_graph}). 
According to the visual results (\cref{fig_comp_style}), models generally reflect the content of the text prompt while capturing the style, such as texture, from the image prompt. However, this tendency varies across models. IP-Adapter and SSR-Encoder fail to adequately incorporate the image prompt, as they balance both prompts by adjusting the proportion of image prompt influence. In contrast, RIVAL and StyleAligned struggle to capture the text prompt, generating unrealistic images due to an overemphasis on the image prompt features rather than the generated features, caused by KV replacement and attention shift, respectively. Our method, however, successfully integrates both text and image prompts by balancing both features across all time steps via Stratified Attention.

\paragraph{Structure-guided Image Generation}
\label{sec_exp_controlnet}
We assess the effectiveness of our method compared to baselines in reflecting the image prompt when additional structural conditions are given. Leveraging ControlNet~\cite{zhang2023adding} to integrate structural conditions into generated images, we evaluate performance both quantitatively and visually under two specific conditions: soft edges and depth. 
For assessing structural faithfulness of the generated image, we measure the DINO similarity (DINO-C) between the structural condition and the generated image.
Quantitative results reveal that our approach is competitive with baselines in terms of structural condition alignment while achieving the most faithful incorporation of both prompts (\cref{table_controlnet}). 
Visual comparisons further support this trend (\cref{fig_comp_cn_depth} and \cref{fig_comp_cn_softedge}).

\subsection{Effects on the Alignment of Image Prompt}
\label{sec_analysis}
\begin{figure}[t!]
    \centering
    \includegraphics[width=\linewidth]{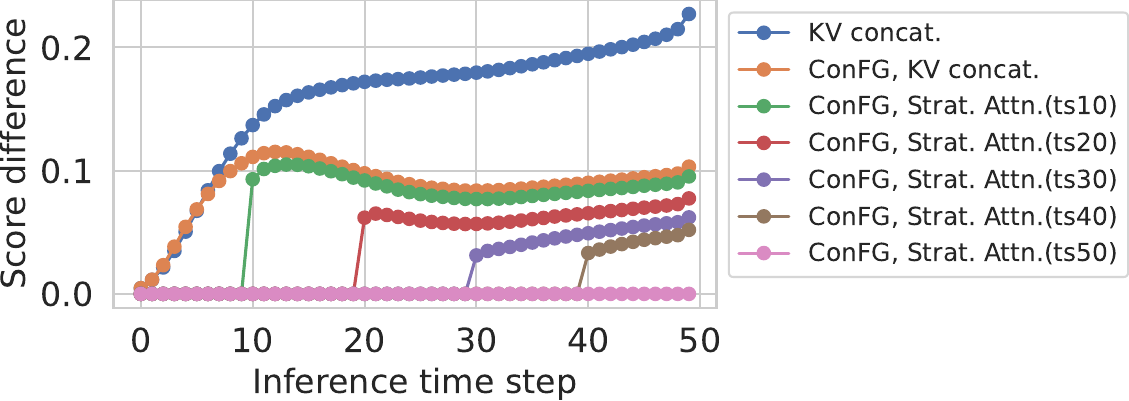}
    \caption{Impact of conflict-free guidance (ConFG) in the self-attention modification and Stratified Attention (Strat. Attn.) on the attention score of the image prompt. A smaller difference (y-axis) indicates greater attention to the image prompt, reflecting it more faithfully in the generated images. Both conflict-free guidance and using longer time steps with Stratified Attention enhance the image prompt alignment.}
    \label{fig_scorediff}
\end{figure}

By investigating the attention scores associated with the image prompts in \cref{eq_attn_stratattn}, we assess the influence of the two primary components of our method on the alignment of the image prompt: 1) conflict-free guidance (ConFG) in~\cref{sec_rm_negprompt} and 2) Stratified Attention (\cref{sec_stratattn}).
Specifically, we compute the difference of the self-attention scores between generated image and the image prompt, where smaller differences indicates that the image prompt is effectively reflected in the generation process.
We compare the differences across diffusion time steps, for KV concatenation (concat.), ConFG, and ConFG with Stratified Attention (Strat. Attn.).
As shown in \cref{fig_scorediff}, conflict-free guidance--by excluding the image prompt as a negative prompt--significantly improves alignment with the image prompt. Additionally, extending the time steps with Stratified Attention further amplifies this effect. These findings validate our hypothesis that both techniques substantially enhance the alignment with the image prompt.

\subsection{Generalization Across Diffusion Models}
\label{sec_sdversion}
\begin{figure}[t!]
    \centering
    \includegraphics[width=\linewidth]{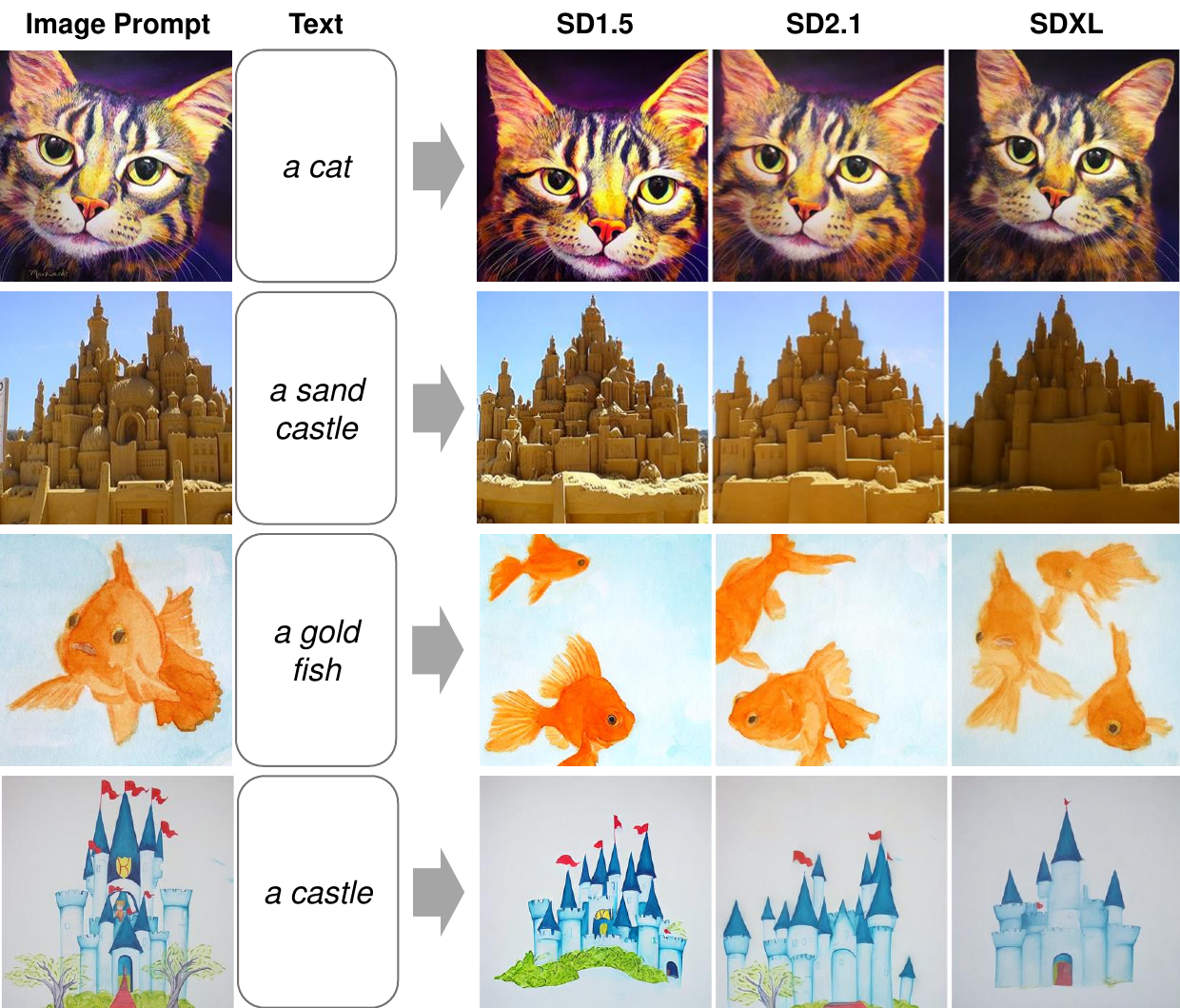}
    \caption{Qualitative comparison of image variation performance across three different diffusion models shows that our method generates images that faithfully reflect the image prompt, regardless of the underlying diffusion model.} 
    \label{fig_imgvar_sd}
\end{figure}

To validate the generalization efficacy of our proposed model, we conduct both qualitative (\cref{fig_imgvar_sd}) and quantitative (\cref{table_imgvar_sd}) comparisons of its performance across various diffusion models, including Stable Diffusion~\cite{rombach2022high} version 1.5, version 2.1, and Stable Diffusion XL~\cite{podell2023sdxl}. Both comparisons demonstrate the applicability of our method across different models. Despite being training-free, our method outperforms the training-based model in reflecting image prompts while remaining competitive across other metrics (\cref{table_imgvar_sd}).

\begin{table}[t!]
\centering
\resizebox{\linewidth}{!}{%
\begin{tabular}{@{}c|cc|cc|cc@{}}
\toprule
 & \multicolumn{2}{c|}{CLIP-I$\uparrow$} & \multicolumn{2}{c|}{CLIP-T$\uparrow$} & \multicolumn{2}{c}{DINO-G$\uparrow$} \\ \midrule
 & Ours & IP-Adapter & Ours & IP-Adapter & Ours & IP-Adapter \\ \midrule
SD1.5 & 0.883 & 0.870 & 0.311 & 0.300 & 0.142 & 0.092 \\
SD2.1 & 0.893 & - & 0.311 & - & 0.118 & - \\
SDXL & 0.865 & 0.859 & 0.298 & 0.299 & 0.110 & 0.121 \\ \bottomrule
\end{tabular}%
}
\caption{Our method demonstrates competitive performance across various diffusion models and outperforms the training-based model~\cite{ye2023ip}, even without the need for training.}
\label{table_imgvar_sd}
\end{table}

\section{Conclusion}
\label{sec_conclusion}

We propose a novel image-prompting method to generate images that faithfully reflect the details of user-provided image conditions. The key idea behind our approach is to eliminate the undesired role of the image prompt as a negative condition during classifier-free guidance in self-attention modification, thereby providing the model with a clearer signal through conflict-free guidance to incorporate the image prompt. Additionally, we introduce Stratified Attention, an alternative self-attention modification that enhances the incorporation of the image prompt by providing a balanced attention flow between the generated image and the image prompt.

We show the superiority of our approach over existing image-prompting models by conducting extensive evaluations across three distinct image generation tasks: image variation, cross-prompt image generation, and structure-guided image generation. 
In addition to these tasks, we aim to explore the scalability and adaptability of our method to various tasks, such as story visualization and style transfer, as part of our future work.

Our work also has limitations. Since this method uses a text-to-image diffusion model in a training-free manner, the quality of generated images depends on the diffusion model's performance. Consequently, issues such as poor hand generation quality persist, but we expect these to improve as diffusion models advance.

\paragraph{Acknowledgement}
{This work was supported by the National Research Foundation of Korea(NRF) grant funded by the Korea government(MSIT) (No. RS-2025-00555621) and Institute for Information \& communications Technology Planning \& Evaluation(IITP) grant funded by the Korea government(MSIT) (RS-2019-II190075, Artificial Intelligence Graduate School Program(KAIST)).}

{
    \small
    \bibliographystyle{ieeenat_fullname}
    \bibliography{main}

\begin{thebibliography}{23}
\providecommand{\natexlab}[1]{#1}
\providecommand{\url}[1]{\texttt{#1}}
\expandafter\ifx\csname urlstyle\endcsname\relax
  \providecommand{\doi}[1]{doi: #1}\else
  \providecommand{\doi}{doi: \begingroup \urlstyle{rm}\Url}\fi

\bibitem[Cao et~al.(2023)Cao, Wang, Qi, Shan, Qie, and Zheng]{cao2023masactrl}
Mingdeng Cao, Xintao Wang, Zhongang Qi, Ying Shan, Xiaohu Qie, and Yinqiang Zheng.
\newblock Masactrl: Tuning-free mutual self-attention control for consistent image synthesis and editing.
\newblock In \emph{ICCV}, pages 22560--22570, 2023.

\bibitem[Caron et~al.(2021)Caron, Touvron, Misra, J{\'e}gou, Mairal, Bojanowski, and Joulin]{caron2021emerging}
Mathilde Caron, Hugo Touvron, Ishan Misra, Herv{\'e} J{\'e}gou, Julien Mairal, Piotr Bojanowski, and Armand Joulin.
\newblock Emerging properties in self-supervised vision transformers.
\newblock In \emph{ICCV}, pages 9650--9660, 2021.

\bibitem[Chung et~al.(2024)Chung, Hyun, and Heo]{chung2024style}
Jiwoo Chung, Sangeek Hyun, and Jae-Pil Heo.
\newblock Style injection in diffusion: A training-free approach for adapting large-scale diffusion models for style transfer.
\newblock In \emph{CVPR}, pages 8795--8805, 2024.

\bibitem[Hertz et~al.(2024)Hertz, Voynov, Fruchter, and Cohen-Or]{hertz2024style}
Amir Hertz, Andrey Voynov, Shlomi Fruchter, and Daniel Cohen-Or.
\newblock Style aligned image generation via shared attention.
\newblock In \emph{CVPR}, pages 4775--4785, 2024.

\bibitem[Heusel et~al.(2017)Heusel, Ramsauer, Unterthiner, Nessler, and Hochreiter]{heusel2017gans}
Martin Heusel, Hubert Ramsauer, Thomas Unterthiner, Bernhard Nessler, and Sepp Hochreiter.
\newblock Gans trained by a two time-scale update rule converge to a local nash equilibrium.
\newblock \emph{NeurIPS}, 30, 2017.

\bibitem[Ho and Salimans(2021)]{ho2021classifier}
Jonathan Ho and Tim Salimans.
\newblock Classifier-free diffusion guidance.
\newblock In \emph{NeurIPS 2021 Workshop on Deep Generative Models and Downstream Applications}, 2021.

\bibitem[Karras(2017)]{karras2017progressive}
Tero Karras.
\newblock Progressive growing of gans for improved quality, stability, and variation.
\newblock \emph{arXiv preprint arXiv:1710.10196}, 2017.

\bibitem[Li et~al.(2022)Li, Li, Xiong, and Hoi]{li2022blip}
Junnan Li, Dongxu Li, Caiming Xiong, and Steven Hoi.
\newblock Blip: Bootstrapping language-image pre-training for unified vision-language understanding and generation.
\newblock In \emph{ICML}, pages 12888--12900. PMLR, 2022.

\bibitem[Lin et~al.(2014)Lin, Maire, Belongie, Hays, Perona, Ramanan, Doll{\'a}r, and Zitnick]{lin2014microsoft}
Tsung-Yi Lin, Michael Maire, Serge Belongie, James Hays, Pietro Perona, Deva Ramanan, Piotr Doll{\'a}r, and C~Lawrence Zitnick.
\newblock Microsoft coco: Common objects in context.
\newblock In \emph{ECCV}, pages 740--755. Springer, 2014.

\bibitem[Liu et~al.(2024)Liu, Wang, Cao, Jia, and Huang]{liu2024towards}
Bingyan Liu, Chengyu Wang, Tingfeng Cao, Kui Jia, and Jun Huang.
\newblock Towards understanding cross and self-attention in stable diffusion for text-guided image editing.
\newblock In \emph{CVPR}, pages 7817--7826, 2024.

\bibitem[Podell et~al.(2023)Podell, English, Lacey, Blattmann, Dockhorn, M{\"u}ller, Penna, and Rombach]{podell2023sdxl}
Dustin Podell, Zion English, Kyle Lacey, Andreas Blattmann, Tim Dockhorn, Jonas M{\"u}ller, Joe Penna, and Robin Rombach.
\newblock Sdxl: Improving latent diffusion models for high-resolution image synthesis.
\newblock \emph{arXiv preprint arXiv:2307.01952}, 2023.

\bibitem[Radford et~al.(2021)Radford, Kim, Hallacy, Ramesh, Goh, Agarwal, Sastry, Askell, Mishkin, Clark, et~al.]{radford2021learning}
Alec Radford, Jong~Wook Kim, Chris Hallacy, Aditya Ramesh, Gabriel Goh, Sandhini Agarwal, Girish Sastry, Amanda Askell, Pamela Mishkin, Jack Clark, et~al.
\newblock Learning transferable visual models from natural language supervision.
\newblock In \emph{ICML}, pages 8748--8763. PMLR, 2021.

\bibitem[Rombach et~al.(2022)Rombach, Blattmann, Lorenz, Esser, and Ommer]{rombach2022high}
Robin Rombach, Andreas Blattmann, Dominik Lorenz, Patrick Esser, and Bj{\"o}rn Ommer.
\newblock High-resolution image synthesis with latent diffusion models.
\newblock In \emph{CVPR}, pages 10684--10695, 2022.

\bibitem[Ruiz et~al.(2023)Ruiz, Li, Jampani, Pritch, Rubinstein, and Aberman]{ruiz2023dreambooth}
Nataniel Ruiz, Yuanzhen Li, Varun Jampani, Yael Pritch, Michael Rubinstein, and Kfir Aberman.
\newblock Dreambooth: Fine tuning text-to-image diffusion models for subject-driven generation.
\newblock In \emph{Proceedings of the IEEE/CVF conference on computer vision and pattern recognition}, pages 22500--22510, 2023.

\bibitem[Song et~al.(2020)Song, Meng, and Ermon]{song2020denoising}
Jiaming Song, Chenlin Meng, and Stefano Ermon.
\newblock Denoising diffusion implicit models.
\newblock \emph{arXiv preprint arXiv:2010.02502}, 2020.

\bibitem[Tumanyan et~al.(2023)Tumanyan, Geyer, Bagon, and Dekel]{tumanyan2023plug}
Narek Tumanyan, Michal Geyer, Shai Bagon, and Tali Dekel.
\newblock Plug-and-play diffusion features for text-driven image-to-image translation.
\newblock In \emph{CVPR}, pages 1921--1930, 2023.

\bibitem[Wang et~al.(2024)Wang, Bai, Wang, Qin, Chen, Li, Tang, and Hu]{wang2024instantid}
Qixun Wang, Xu Bai, Haofan Wang, Zekui Qin, Anthony Chen, Huaxia Li, Xu Tang, and Yao Hu.
\newblock Instantid: Zero-shot identity-preserving generation in seconds.
\newblock \emph{arXiv preprint arXiv:2401.07519}, 2024.

\bibitem[Xu et~al.(2024)Xu, Guo, Wang, Huang, Essa, and Shi]{xu2024prompt}
Xingqian Xu, Jiayi Guo, Zhangyang Wang, Gao Huang, Irfan Essa, and Humphrey Shi.
\newblock Prompt-free diffusion: Taking" text" out of text-to-image diffusion models.
\newblock In \emph{CVPR}, pages 8682--8692, 2024.

\bibitem[Ye et~al.(2023)Ye, Zhang, Liu, Han, and Yang]{ye2023ip}
Hu Ye, Jun Zhang, Sibo Liu, Xiao Han, and Wei Yang.
\newblock Ip-adapter: Text compatible image prompt adapter for text-to-image diffusion models.
\newblock \emph{arXiv preprint arXiv:2308.06721}, 2023.

\bibitem[Zhang et~al.(2023)Zhang, Rao, and Agrawala]{zhang2023adding}
Lvmin Zhang, Anyi Rao, and Maneesh Agrawala.
\newblock Adding conditional control to text-to-image diffusion models.
\newblock In \emph{ICCV}, pages 3836--3847, 2023.

\bibitem[Zhang et~al.(2024{\natexlab{a}})Zhang, Song, Liu, Wang, Yu, Tang, Li, Tang, Hu, Pan, et~al.]{zhang2024ssr}
Yuxuan Zhang, Yiren Song, Jiaming Liu, Rui Wang, Jinpeng Yu, Hao Tang, Huaxia Li, Xu Tang, Yao Hu, Han Pan, et~al.
\newblock Ssr-encoder: Encoding selective subject representation for subject-driven generation.
\newblock In \emph{CVPR}, pages 8069--8078, 2024{\natexlab{a}}.

\bibitem[Zhang et~al.(2024{\natexlab{b}})Zhang, Xing, Lo, and Jia]{zhang2024real}
Yuechen Zhang, Jinbo Xing, Eric Lo, and Jiaya Jia.
\newblock Real-world image variation by aligning diffusion inversion chain.
\newblock \emph{NeurIPS}, 36, 2024{\natexlab{b}}.

\bibitem[Zhou et~al.(2024)Zhou, Zhou, Cheng, Feng, and Hou]{zhou2024storydiffusion}
Yupeng Zhou, Daquan Zhou, Ming-Ming Cheng, Jiashi Feng, and Qibin Hou.
\newblock Storydiffusion: Consistent self-attention for long-range image and video generation.
\newblock \emph{arXiv preprint arXiv:2405.01434}, 2024.

\end{thebibliography}
}

\clearpage
\setcounter{page}{1}
\setcounter{section}{0}
\setcounter{figure}{0}
\setcounter{table}{0}
\renewcommand\thesection{\Alph{section}}
\renewcommand{\thefigure}{\Alph{figure}}
\renewcommand{\thetable}{\Alph{table}}

\maketitlesupplementary
\section{Overview}
In this supplementary, we first provide the implementation details of the proposed method (\cref{sec_imp_detail}). Next, expanding the analysis in \cref{sec_analysis}, we further analyze the impact of the proposed method on alignment with the image prompt across three aspects: self-attention layers, diffusion models, and datasets (\cref{sec_supple_analysis}). Subsequently, we conduct ablation studies (\cref{sec_ablation}) and analyze the impact of the negative image prompt and weighted attention in the proposed conflict-free guidance (\cref{sec_neg_ip}) and Stratified Attention (\cref{sec_weight_stratattn}), respectively, to demonstrate the method’s effectiveness in faithfully reflecting the given image prompt and its details in the generated images. In addition, we explain the role of text and image prompts in our model (\cref{sec_role_textprompt}) and compare the performance in image prompt alignment with identity-preserving methods (\cref{sec_comp_identity}). Finally, we highlight the superiority of the proposed method over existing approaches as well as its potential for cross-prompt image generation through additional qualitative results (\cref{sec_add_qual}).

\section{Implementation Details}
\label{sec_imp_detail}
We conduct all experiments with 50 inference steps for each diffusion model. For the classifier-free guidance scale, we set the value to 7.5 for Stable Diffusion (SD)~\cite{rombach2022high} and 5.0 for Stable Diffusion XL~\cite{podell2023sdxl}. Additionally, the self-attention modification is applied exclusively to the decoder blocks, and Stratified Attention performs a weighted sum of the attention for the generated image and the image prompt, assigning weights of $\lambda_G = 0.5$ and $\lambda_R = 0.5$, respectively.
We also employ Stratified Attention up to 10 time steps, except for structure-guided image generation using ControlNet, where it is applied up to 25 time steps.
The positive text prompt is appended with ``, best quality, extremely detailed.'' at the end of the sentence, while the negative text prompt is set to ``monochrome, bad anatomy, bad hands, cropped, worst quality.''. To ensure a fair comparison and eliminate the influence of text prompts, we use the same text prompt across all models, including the baselines.
Following RIVAL's setting, we use null prompts for DDIM inversion in cross-prompt image generation, while employing positive text prompts in all other experiments.

\begin{figure}[t!]
    \centering
    \begin{subfigure}[b]{\linewidth}
        \centering
        \includegraphics[width=\linewidth]{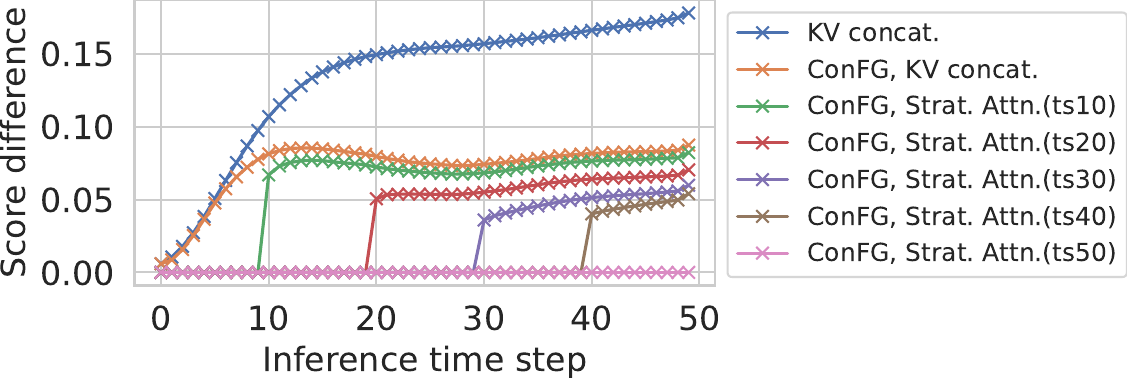}
        \caption{Diffusion model: SD v2.1}
        \label{fig_scorediff_sd21}
    \end{subfigure}
    
    \vspace{5mm}
    \begin{subfigure}[b]{\linewidth}
        \centering
        \includegraphics[width=\linewidth]{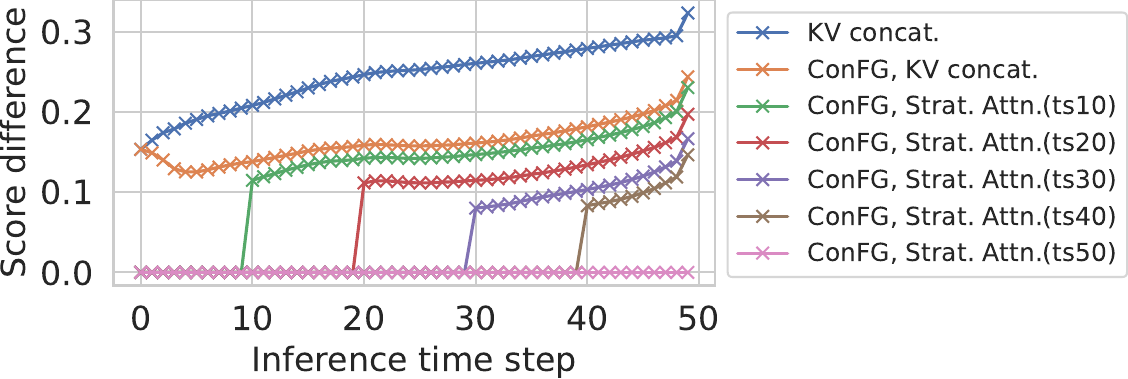}
        \caption{Diffusion model: SDXL}
        \label{fig_scorediff_sdxl}
    \end{subfigure}

    \vspace{5mm}
    \begin{subfigure}[b]{\linewidth}
        \centering
        \includegraphics[width=\linewidth]{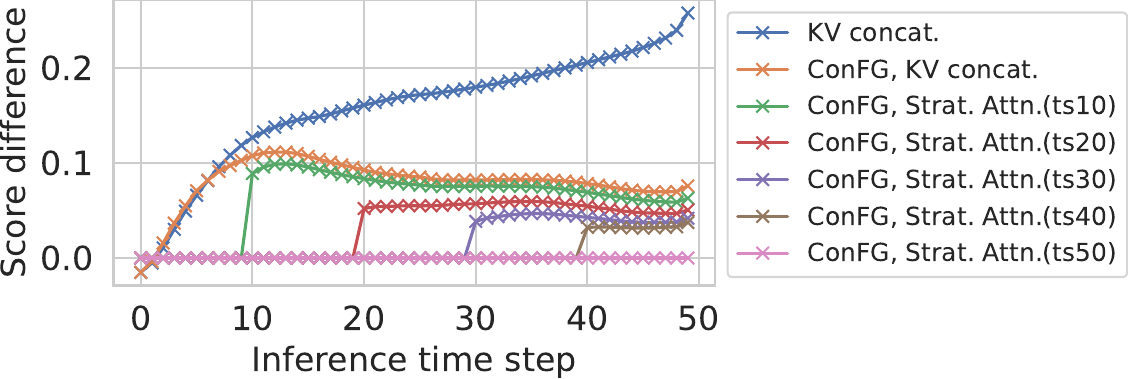}
        \caption{Dataset: ImageNet-R-TI2I}
        \label{fig_scorediff_imgnet}
    \end{subfigure}
    
    \vspace{5mm}
    \begin{subfigure}[b]{\linewidth}
        \centering
        \includegraphics[width=\linewidth]{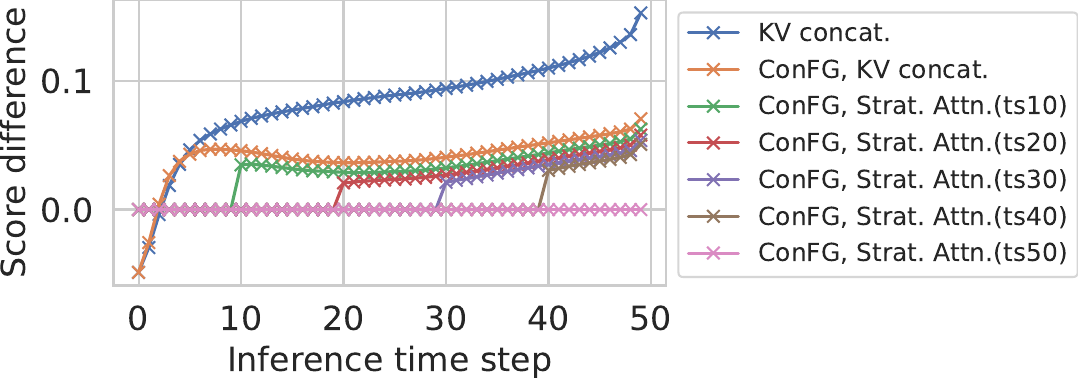}
        \caption{Dataset: CelebAHQ}
        \label{fig_scorediff_celeba}
    \end{subfigure}
    
    \caption{Comparison of the impact of conflict-free guidance (ConFG) and Stratified Attention (Strat. Attn.) on the attention scores of the image prompt across \textit{diffusion models} (a and b) and \textit{datasets} (c and d). A smaller difference on the y-axis indicates stronger attention to the image prompt, resulting in a more faithful alignment in the generated images.}
    \vspace{-3mm}
    \label{fig_scorediff_supple}
\end{figure}

\begin{figure*}[ht!]
    \centering
    \includegraphics[width=\textwidth]{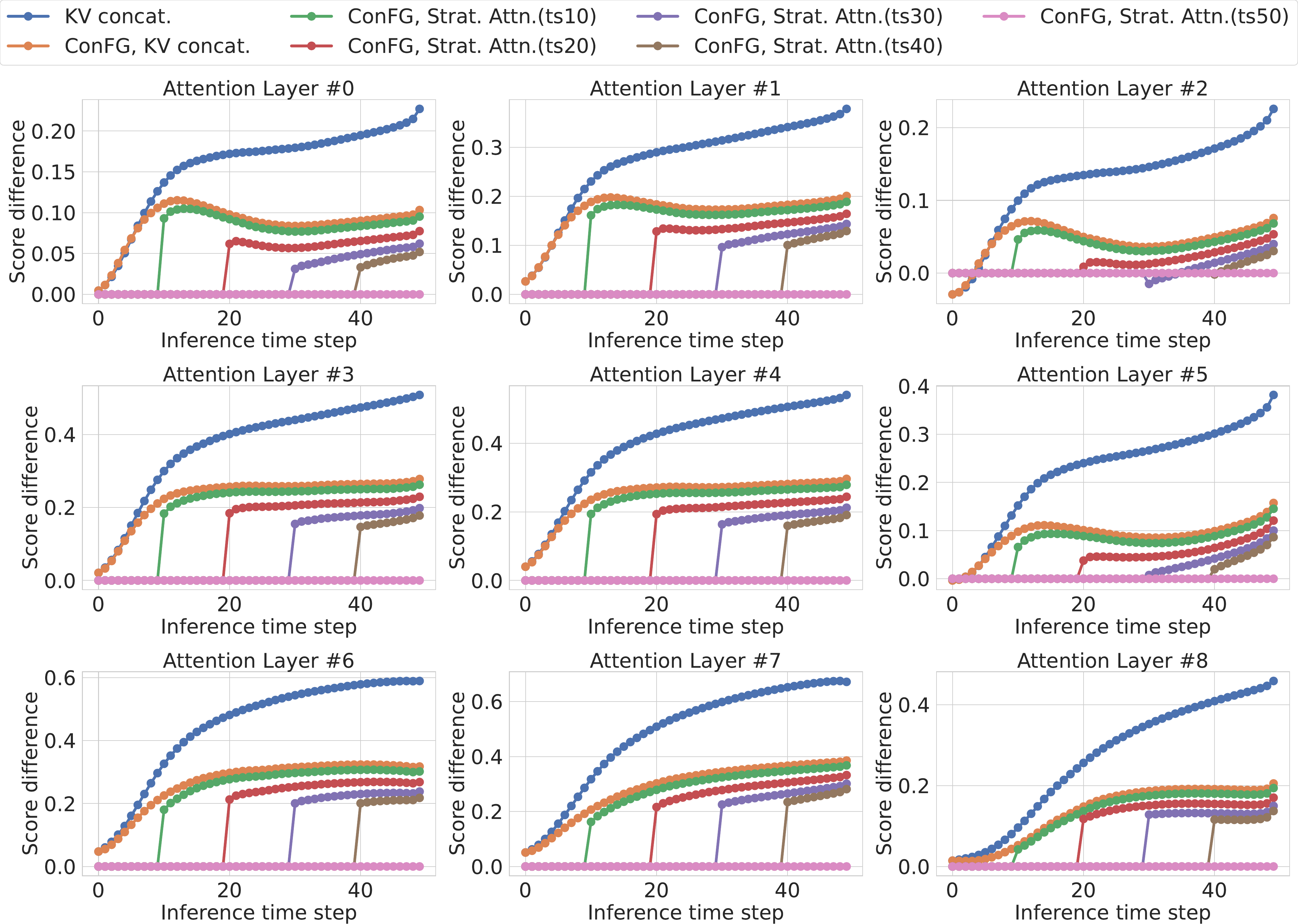}
    \caption{Comparison of the impact of conflict-free guidance (ConFG) and Stratified Attention (Strat. Attn.) on the attention scores of the image prompt across \textit{self-attention layers}. A smaller difference on the y-axis indicates stronger attention to the image prompt, resulting in a more faithful alignment in the generated images.}
    \label{fig_scorediff_layer}
\end{figure*}

\section{Effects on the Alignment of Image Prompt}
\label{sec_supple_analysis}
In \cref{sec_analysis}, we demonstrate the effectiveness of two key components, conflict-free guidance and Stratified Attention, of the proposed method in increasing the attention score of the image prompt by visualizing changes in attention scores within the self-attention layer of Stable Diffusion, using 100 randomly sampled images from the COCO validation set (\cref{fig_scorediff}). However, this analysis is limited to results from the first of nine self-attention layers in the decoder in Stable Diffusion version 1.5 and relies on a single dataset. In this section, we broaden the analysis by reporting results from additional layers, diffusion models, and datasets to confirm the consistent performance of the proposed method across these three factors.

First, we present the results for all layers of SD version 1.5, including the first layer, in \cref{fig_scorediff_layer}. While the attention scores of the image prompt show slight variations across layers, they consistently improve as we apply conflict-free guidance (ConFG) or use longer time steps in Stratified Attention (Strat. Attn.). Furthermore, we observe that these trends remain consistent across all layers in the subsequent experiments, even though we visualize the results from the first layer.

Next, we validate that the two proposed components consistently enhance the attention score of the image prompt across different diffusion models by conducting the same experiment with SD version 2.1 (\cref{fig_scorediff_sd21}) and SDXL (\cref{fig_scorediff_sdxl}). The results indicate that, while the attention score values differ slightly from those of SD version 1.5, both models exhibit the same overall trend. Specifically, a minor difference arises at the initial time step: SD models generally produce nearly identical attention scores for the image prompt and the generated features, whereas SDXL yields slightly higher attention scores for the generated features. Despite this variation, the proposed method consistently improves the attention score of the image prompt across all models. These findings confirm that the proposed method reliably increases the attention score of the image prompt, regardless of the diffusion model. Additionally, as shown in \cref{table_imgvar_sd} and \cref{fig_imgvar_sd}, this improvement strengthens alignment with the image prompt, further demonstrating the robustness of the method.

\begin{table*}[t!]
    \centering
    \begin{minipage}{0.43\linewidth}
        \centering
        \resizebox{\linewidth}{!}{%
        \begin{tabular}{@{}c|cc@{}}
        \toprule
         & CLIP-I $\uparrow$ & CLIP-T $\uparrow$ \\ \midrule
        KV concatenation & 0.770 & \textbf{0.316} \\
        ConFG + KV concatenation & 0.882 & 0.311 \\ 
        ConFG + Stratified Attention (ts10) & 0.886 & 0.311 \\
        ConFG + Stratified Attention (ts20) & 0.893 & 0.310 \\
        ConFG + Stratified Attention (ts30) & 0.899 & 0.308 \\
        ConFG + Stratified Attention (ts40) & 0.901 & 0.307 \\
        ConFG + Stratified Attention (ts50) & \textbf{0.902} & 0.306 \\ \bottomrule
        \end{tabular}%
        }
        \caption{Ablation studies for key components: Conflict-free guidance (ConFG) and Stratified Attention. Applying ConFG or increasing the time steps of Stratified Attention enhances the faithful reflection of the image prompt in the generated image.}
        \label{tab_ablation}
    \end{minipage}%
    \hfill
    \begin{minipage}{0.55\linewidth}
        \centering
        \resizebox{\linewidth}{!}{%
        \begin{tabular}{@{}cc|cc@{}}
        \toprule
        $\lambda_P$ (Image prompt) & $\lambda_G$ (Generated fetures) & CLIP-I $\uparrow$ & CLIP-T $\uparrow$ \\ \midrule
        1.00 & 0.00 & \textbf{0.917} & 0.301 \\
        0.67 & 0.33 & 0.908 & 0.303 \\
        0.50 & 0.50 & 0.902 & 0.306 \\
        0.33 & 0.67 & 0.884 & 0.309 \\
        0.00 & 1.00 & 0.724 & \textbf{0.313} \\ \bottomrule
        \end{tabular}%
        }
        \caption{Quantitative impact of the weights applied to the attention of the image prompt and generated features in Stratified Attention on the alignment with the image and text prompts. $\lambda_P$ and $\lambda_G$ represent the weights applied to the image prompt and generated features, respectively. Since we use the diffusion model in a training-free manner, the sum of the two weights is set to 1.}
        \label{tab_ablation_lambda}
    \end{minipage}
\end{table*}

\begin{figure*}[t!]
    \centering
    \begin{minipage}{\linewidth}
        \centering
        \includegraphics[width=\linewidth]{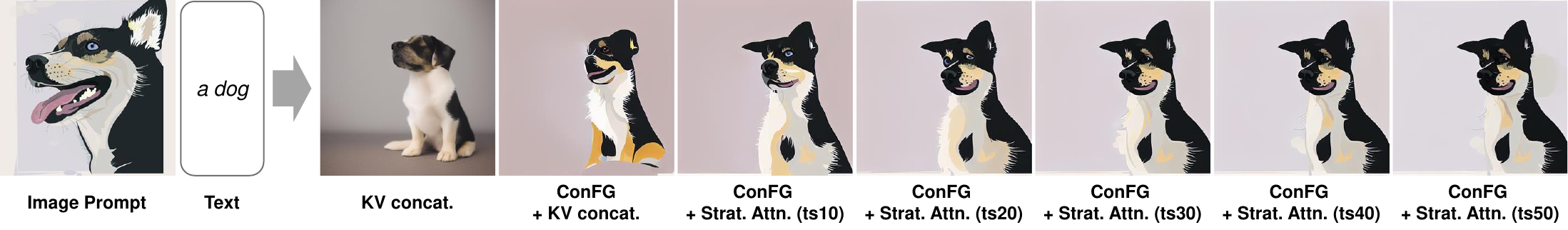}
        \caption{Qualitative results of ablation studies. When Conflict-Free Guidance (ConFG) is not applied, the image prompt serves as both positive and negative guidance, creating conflicts that hinder the generated image from reflecting the given prompt. However, with ConFG, the generated image aligns effectively with the provided prompt, and Stratified Attention (Strat. Attn.) further improves this alignment.}
        \label{fig_ablation}
    \end{minipage}
    
    \vspace{5mm}
    
    \begin{minipage}{\linewidth}
        \centering
        \includegraphics[width=\linewidth]{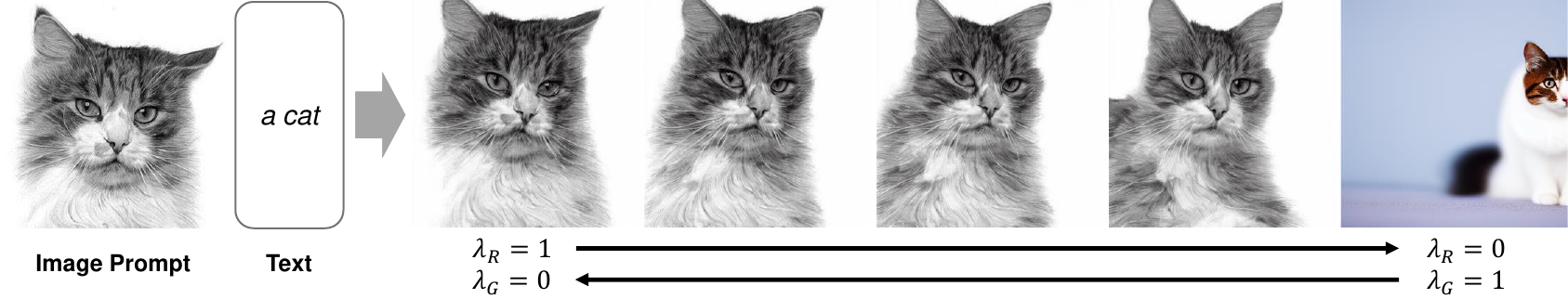}
        \caption{Visual results showing the impact of varying the weights applied to the attention of the image prompt and generated features in Stratified Attention on the generated images. When the weight for the image prompt, $\lambda_P$, is set to 0, the image prompt is not incorporated at all. However, the separated attention score computation of Stratified Attention ensures that even a small weight for the image prompt allows it to be faithfully reflected. Additionally, as $\lambda_P$ increases, the generated images progressively resemble the structure of the image prompt.}
        \label{fig_ablation_lambda}
    \end{minipage}
\end{figure*}

Finally, we confirm that the two key components consistently enhance the attention score of the image prompt across datasets by conducting the same experiments on the ImageNet-R-TI2I dataset~\cite{tumanyan2023plug} (\cref{fig_scorediff_imgnet}) and the CelebAHQ dataset~\cite{karras2017progressive} (\cref{fig_scorediff_celeba}). Consistent with previous results, applying conflict-free guidance or increasing the time steps with Stratified Attention reliably improves the attention score of the image prompt, regardless of the dataset. These findings align with the results presented in \cref{table_imgvar}.

From these analyses, we conclude that the proposed method consistently enhances the attention score of the image prompt, ensuring its faithful incorporation into the generated images across diverse diffusion models and datasets. This conclusion is well-supported by the results presented in \cref{sec_exp_results}.

\section{Ablation Study}
\label{sec_ablation}
This section presents ablation studies on the two key components of the proposed method: conflict-free guidance (ConFG) and Stratified Attention (Strat. Attn.). 
To assess the effectiveness of these components in faithfully incorporating the image prompt into the generated images, we generate 20 images per image prompt from the ImageNet-R-TI2I dataset.
Since the proposed method employs an Attention Fusion strategy~\cite{zhang2024real} that combines KV concatenation with Stratified Attention, its performance is compared to KV concatenation (concat.) as the baseline.

First, we quantitatively demonstrate the advantages of the two proposed components in reflecting the image prompt.
As shown in~\cref{tab_ablation}, the quantitative results reveal that the conflict-free guidance slightly decreases the alignment with the text prompt but significantly enhances the incorporation of the image prompt by removing its role as negative guidance. Furthermore, this improvement becomes more evident as the time steps utilizing Stratified Attention increase. 
However, since applying Stratified Attention across all time steps reduce the alignment with text, we recommend using Stratified Attention for time steps between 10 and 30.

Also, the qualitative analysis further supports these findings (\cref{fig_ablation}). Without conflict-free guidance, the image prompt simultaneously acts as both positive and negative guidance, hindering its incorporation into the generated images. In contrast, conflict-free guidance resolves these conflicting roles, allowing the generated images to faithfully reflect the image prompt, including its color tones and textures. Additionally, increasing the time steps with Stratified Attention further improves the ability of the generated images to faithfully reflect the image prompt, including its color and content. 

\begin{figure}[t]
    \centering
    \includegraphics[width=\linewidth]{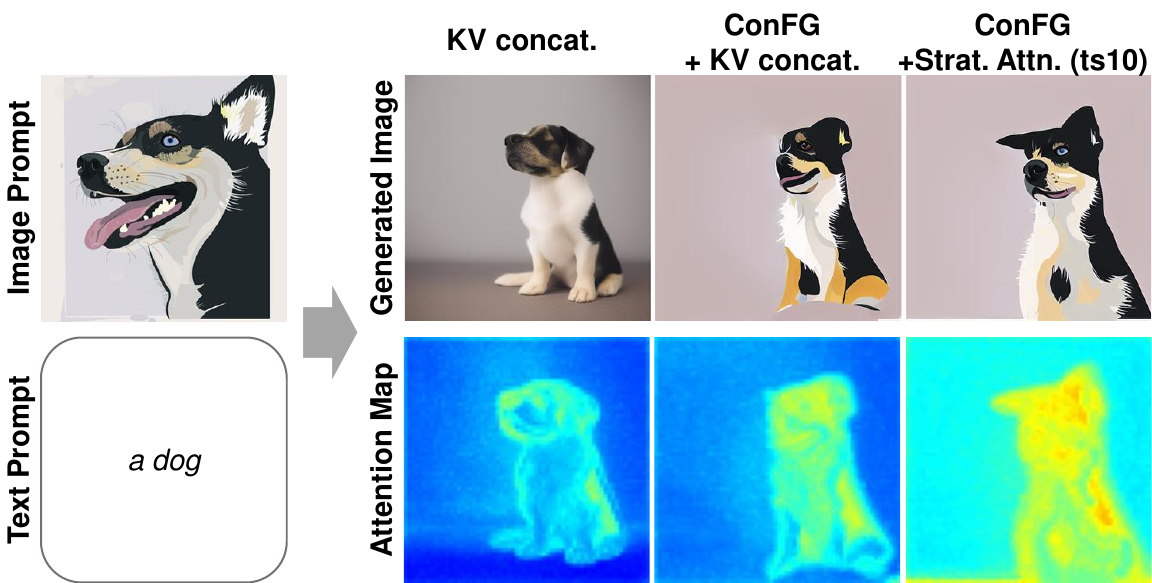}
    \caption{Visualization of the attention map to show the effects of conflict-free guidance (ConFG) and Stratified Attention (Strat. Attn.) on image prompt alignment. The color scale of the attention maps ranges from red, indicating high alignment with the image prompt, to blue, indicating low alignment.}
    \label{fig_reb_attnmap}
\end{figure}

For further analysis, in \cref{fig_reb_attnmap}, we visualize the self-attention maps of the results from KV concatenation, conflict-free guidance, and Stratified Attention presented in \cref{fig_ablation}. According to these results, conflict-free guidance generally enhances the alignment of the image prompt with the subject, while Stratified Attention improves alignment across the entire generated image, including the subject.

From these results, we affirm the crucial role of conflict-free guidance and Stratified Attention in achieving faithful integration of the image prompt into the generated images, thereby demonstrating the effectiveness of the proposed method. 

\section{Impact of Negative Image Prompt on Conflict-free Guidance}
\label{sec_neg_ip}
\begin{figure}[t]    
    \centering
    \includegraphics[width=\linewidth]{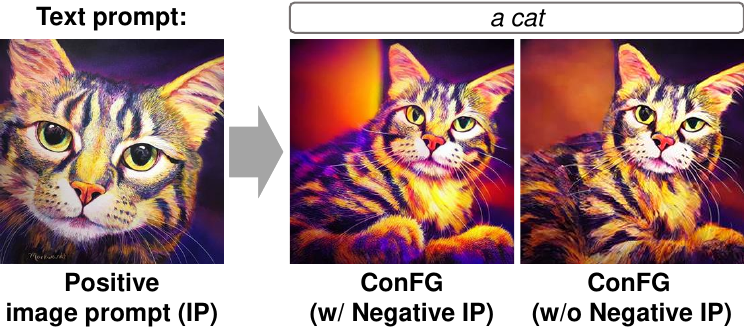}
    \caption{Qualitative results showing the impact of using and not using negative image prompts. Using negative image prompts introduces unintended effects, such as shifts in color tone, in the generated images.}
    \label{fig_reb_diffneg}
\end{figure}

Conflict-free guidance can be implemented in two ways: one where no image is supplied to the negative branch, and the other where an image dissimilar to the provided image prompt is used for the negative branch. To assess the efficacy of our proposed approach, we conduct experiments comparing image prompt alignment under both conditions. In these experiments, a negative image, generated from a negative text prompt, was used (the first image in Figure~\cref{fig_analysis_sacontrol}-a). The results yielded a CLIP-I score of 0.854 and a CLIP-T score of 0.319. When compared to the results obtained using identical image prompts for both branches (the first row in Table~\cref{tab_ablation}), it is evident that directing the image prompt solely to the positive branch leads to improved alignment. However, the CLIP-I score remains lower than that of conflict-free guidance, where no image prompt is provided to the negative branch (second row). We attribute this discrepancy to the negative image prompt containing not only undesirable features (\eg, poor anatomical structure) but also other characteristics (\eg, color), which introduce unintended effects (\eg, shifts in color tone) in the generated images (\cref{fig_reb_diffneg}).

\section{Effects of Weighted Attention in Stratified Attention}
\label{sec_weight_stratattn}
In this section, we analyze the influence of the two hyperparameters, $\lambda_P$ and $\lambda_G$, in Stratified Attention on achieving alignment with the image prompt (\cref{eq_attn_stratattn}). As mentioned in~\cref{sec_stratattn}, Stratified Attention tackles the problem of attention bias, where queries disproportionately focus on keys and values derived from the same generated features in KV concatenation, thereby enhancing the integration of the image prompt. Specifically, it separately computes attention scores for the image prompt and the generated features, subsequently combining them through a weighted sum using $\lambda_P$ and $\lambda_G$. Therefore, these parameters individually determine the extent to which the image prompt and the generated features are integrated into the final generated images. In these experiments, to clearly observe the effects of the weights, Stratified Attention is applied to all time steps. 

To evaluate this, we quantitatively and visually assess the incorporation of the image prompt into the generated images by varying the values of the two hyperparameters. As shown in \cref{tab_ablation_lambda}, the quantitative analysis reveals that increasing $\lambda_P$, the weight assigned to the image prompt, while decreasing $\lambda_G$, the weight assigned to the generated features, enhances the faithful integration of the image prompt into the generated images. Conversely, reversing this adjustment diminishes the influence of the image prompt. The visual results in \cref{fig_ablation_lambda} corroborate this pattern, demonstrating that setting $\lambda_P$ to 0 entirely excludes the image prompt, while progressively increasing $\lambda_P$ incorporates not only the color tones and textures but also the structure of the image prompt. Based on these findings, we recommend setting $\lambda_P$ within the range of 0.3 to 0.7.

\section{Role of Text and Image Prompts}
\label{sec_role_textprompt}
\begin{figure}[t]
    \centering
    \includegraphics[width=\linewidth]{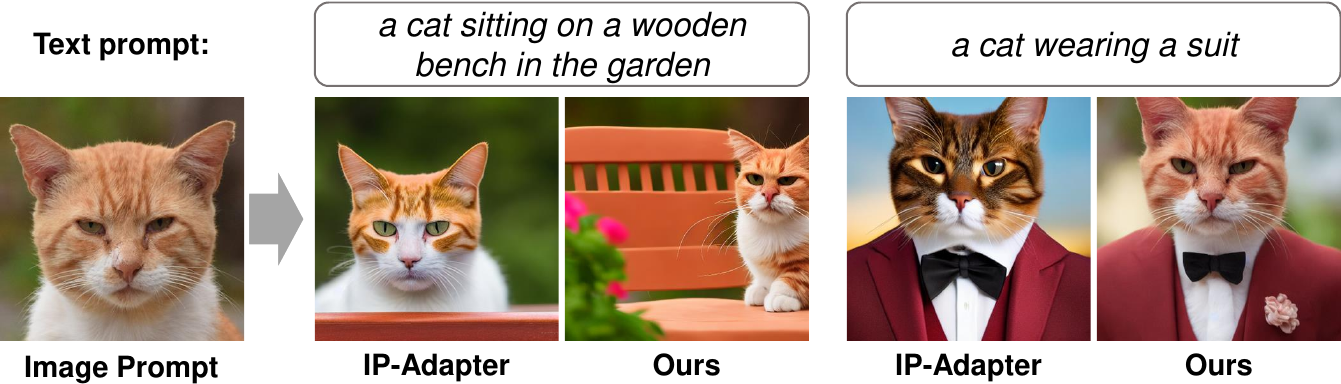}
    \caption{Generated images from realistic scenarios}
    \vspace{-3mm}
    \label{fig_reb_cat}
\end{figure}

Since we use the text-to-image diffusion model in a training-free manner, we input text prompts alongside the image prompt to maintain the model's mechanism. 
In this setup, we typically assume the text and image prompts are aligned and aim to faithfully reflect the details of the image prompt, following the experimental setup of previous studies. However, when the image and text prompts are misaligned (\cref{fig_comp_style}), we can enhance the influence of the text prompt by adjusting hyperparameters, similar to the baselines. As an additional example of misalignment between image and text prompts, we generate images using two text prompts in conjunction with a cat image prompt, and compare the results with those generated by IP-Adapter~\cite{ye2023ip}. This experiment serves to demonstrate the superiority of our method, even in realistic scenarios. As shown in \cref{fig_reb_cat}, our approach produces images that more accurately capture the details of the cat in the image prompt, in contrast to the results from IP-Adapter.

\section{Comparison with Identity-preserving Method}
\label{sec_comp_identity}
We conduct a comparative analysis with identity-preserving approaches~\cite{wang2024instantid}, which generate images while maintaining the identity of a given face, as both models aim to capture the details of the provided image. We use InstantID~\cite{wang2024instantid} as the baseline for identity-preserving methods and evaluate both models using 58 images from the CelebA-HQ dataset~\cite{karras2017progressive}, where InstantID successfully detects faces. The comparison is based on image prompt alignment, and the results demonstrate that our model outperforms InstantID, achieving a CLIP-I score 0.124 higher. This discrepancy underscores the superiority of our method, which not only preserves the identity of the image prompt but also captures additional details, such as texture, while InstantID focuses solely on identity preservation.

\section{Additional Qualitative Results}
\label{sec_add_qual}
In addition to the qualitative results provided in \cref{sec_exp_results}, this section presents additional visual results across three tasks to demonstrate the superiority of our method: cross-prompt image generation with various text prompts, cross-prompt image generation with structural conditions, and image variation. 

\begin{figure*}[t!]
    \centering
    \includegraphics[width=\textwidth]{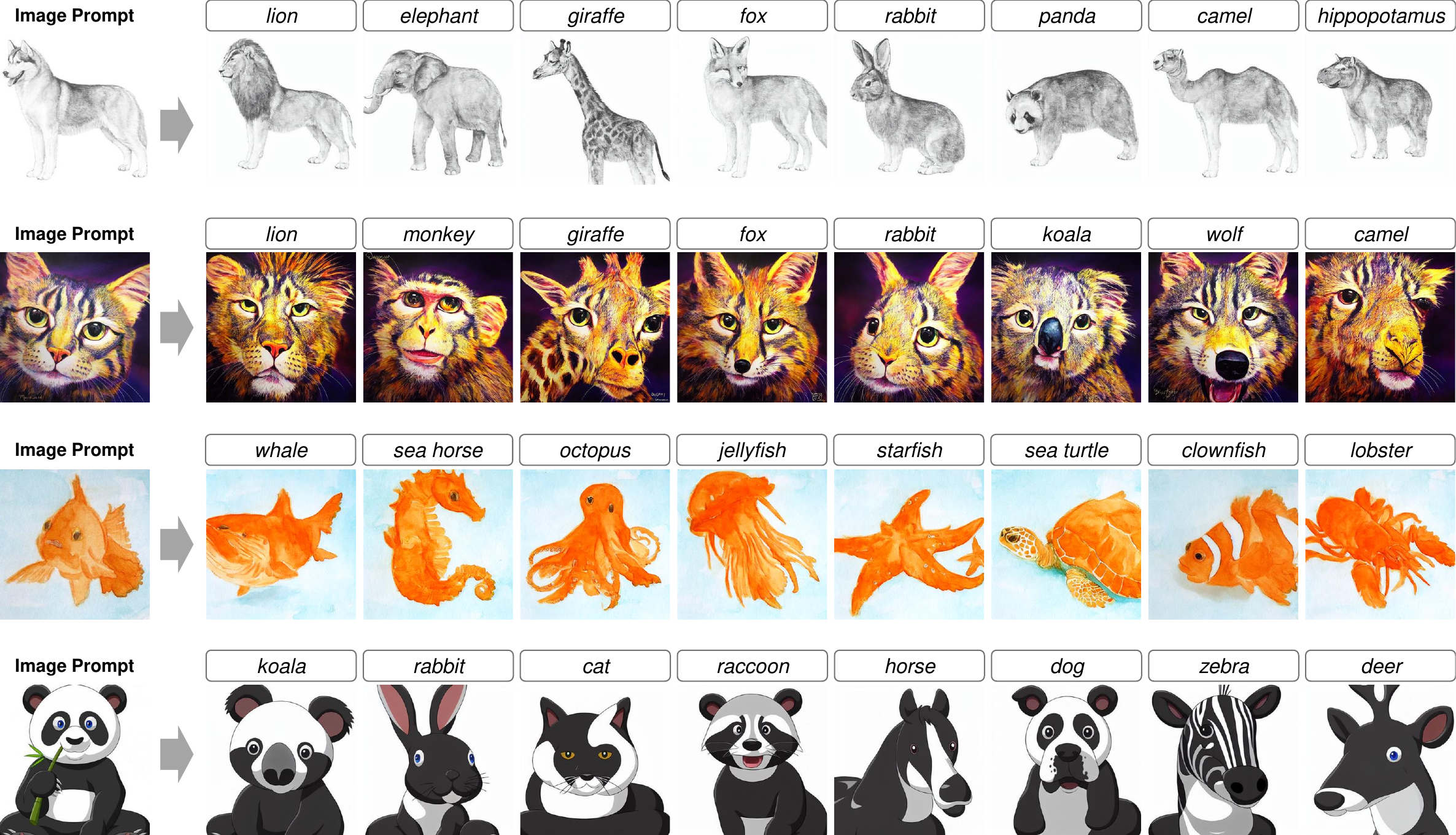}
    \caption{Qualitative results of our method on cross-prompt image generation tasks. The text within the rounded rectangle serves as a text prompt. Our method generates images from text prompts that consistently capture the color tones and textures of the image prompts.}
    \label{fig_supple_crossprompt}
\end{figure*}

First, we present the results of our method in cross-prompt generation tasks with various text prompts. In \cref{sec_exp_style}, we compare its performance to baseline models by generating images from a single image-text pair. Extending this, we pair a single image prompt with multiple text prompts, generating images for each pair to show that our method consistently integrates both prompts. As shown in \cref{fig_supple_crossprompt}, our method reliably reflects the color tones and textures of the image prompt while adapting to diverse text prompts. These results demonstrate that the proposed method enhances controllability in image generation and holds promise for applications such as style transfer.

\begin{figure*}[t!]
    \centering
    \includegraphics[width=\textwidth]{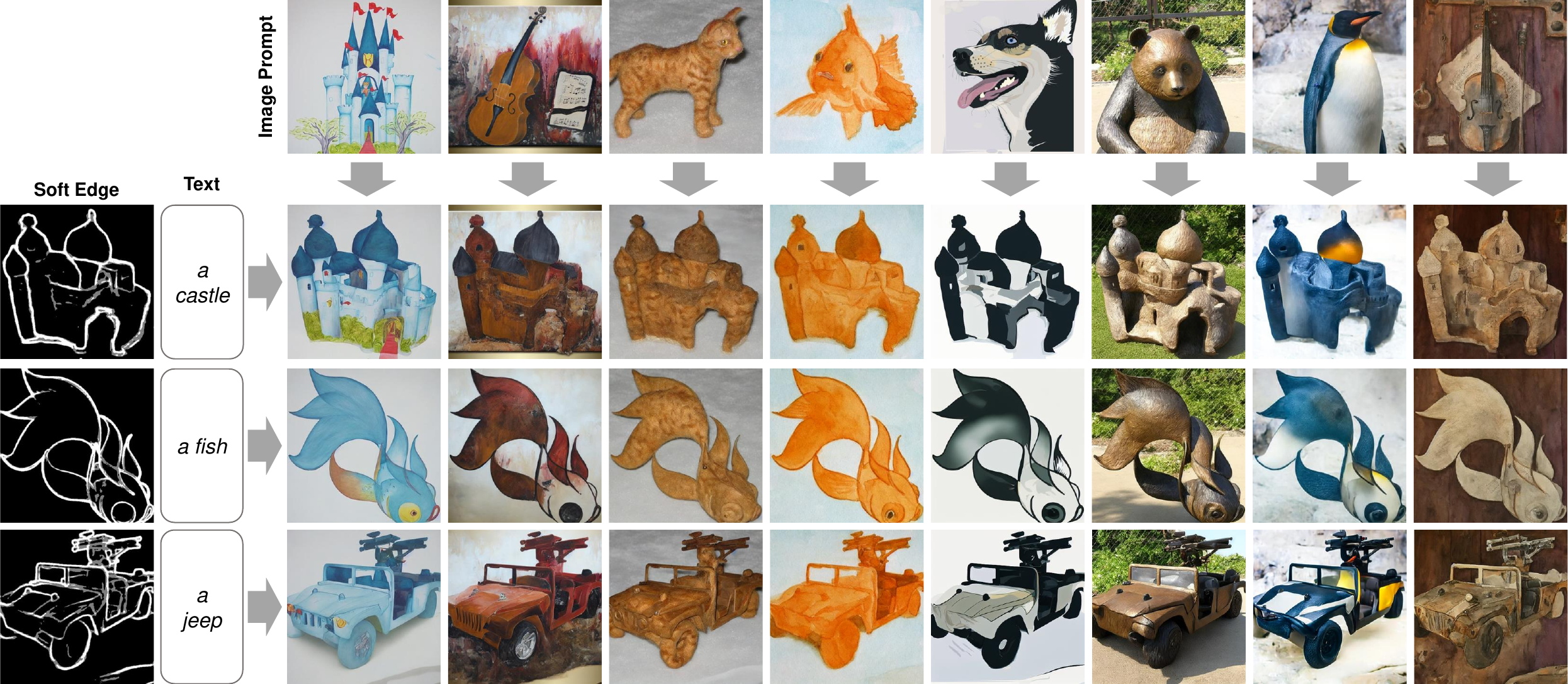}
    \caption{Qualitative results of our method on the combination of cross-prompt image generation and structure-guided image generation tasks. Our method generates images from text prompts that consistently reflect the color tones and textures of the image prompts while capturing the structure of images with soft edges.}
    \label{fig_supple_crossprompt_cn}
\end{figure*}

Next, we demonstrate the performance of the proposed method in cross-prompt image generation with a given structural condition, combining the two tasks presented in \cref{sec_comp_baseline}. To incorporate the structural condition as input, we employ ControlNet~\cite{zhang2023adding}, following the approach used in the main paper. As illustrated in \cref{fig_supple_crossprompt_cn}, the proposed method faithfully integrates the structural condition while generating text-prompted images that consistently reflect the style of the image prompt, including its color tone and texture, as observed in the earlier cross-prompt image generation results. 

\begin{figure*}[t!]
    \centering
    \begin{subfigure}[t]{0.85\textwidth}
        \centering
        \includegraphics[width=\linewidth]{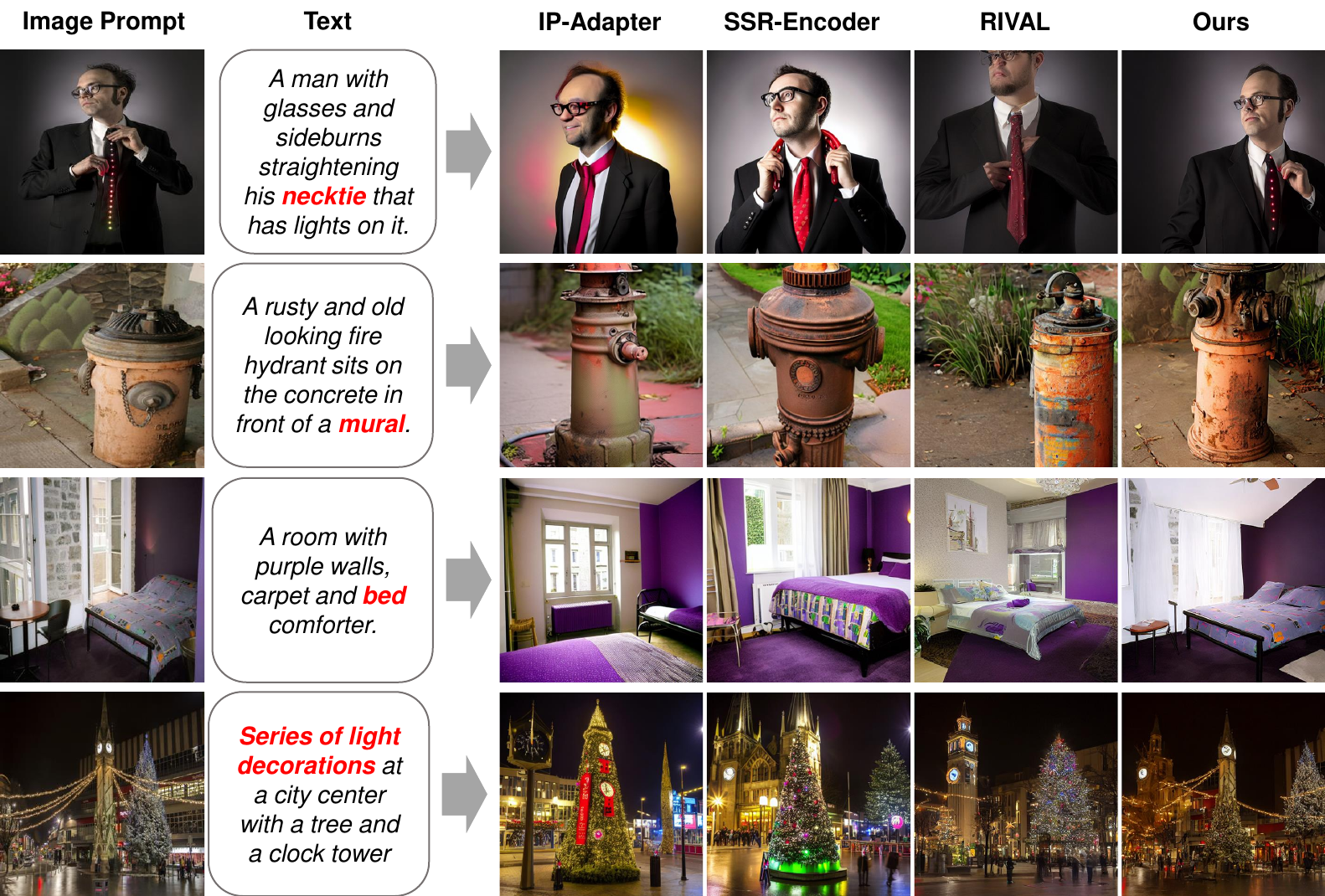}
        \caption{COCO validation. The baselines struggle to reflect the intricate details of the given image prompt, such as the necktie's pattern, the mural, the bed blanket's design, or the light decorations, whereas our method faithfully incorporates these elements.}
        \label{fig_imgvar_cocoval}
    \end{subfigure}
    
    \vspace{1mm}

    \begin{subfigure}[t]{0.85\textwidth}
        \centering
        \includegraphics[width=\linewidth]{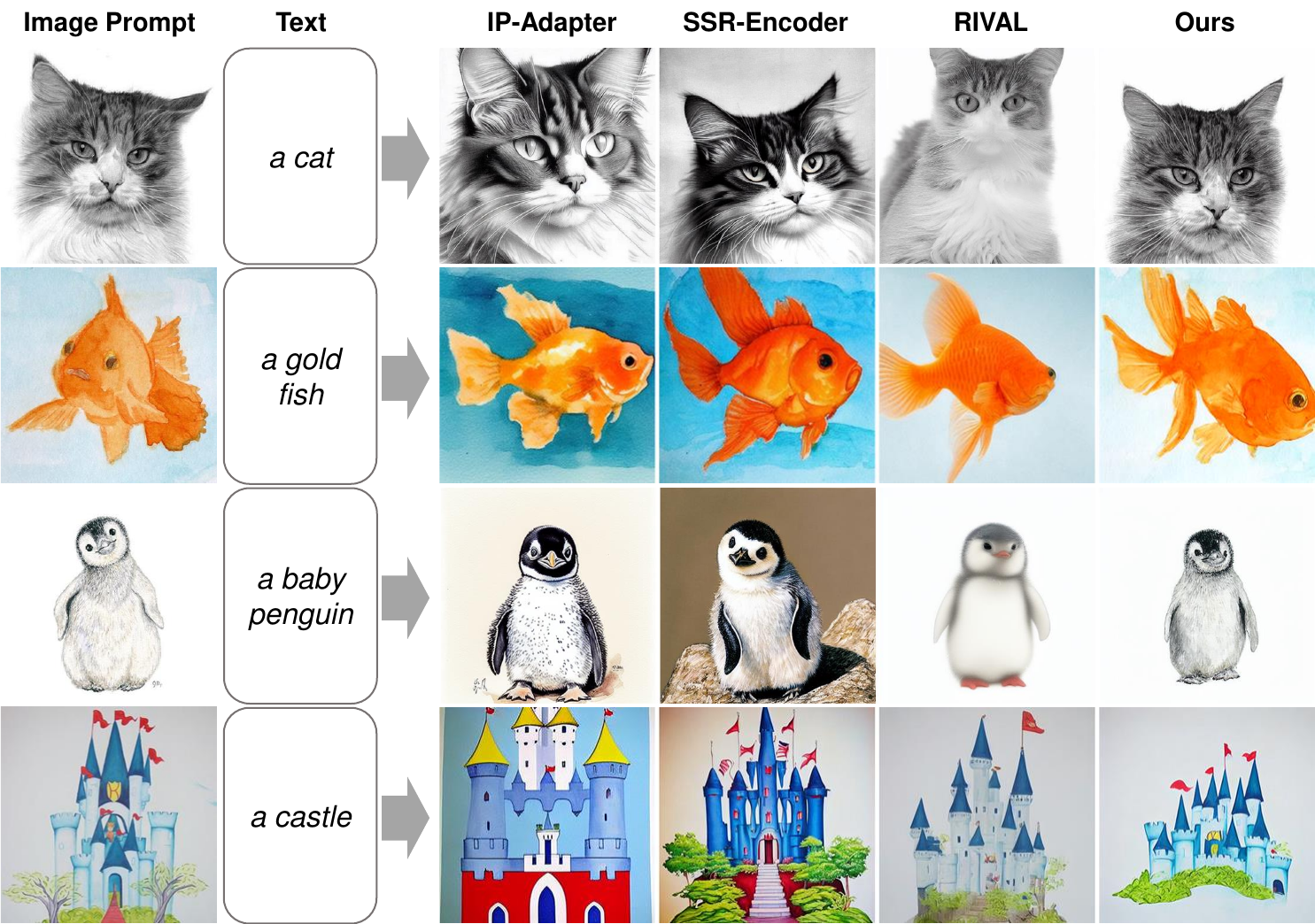}
        \caption{ImageNet-R-I2I. The baselines fail to capture the color tone and texture of the given image prompt, whereas our method faithfully incorporates them.}
        \label{fig_imgvar_imgnetr}
    \end{subfigure}

    \caption{Qualitative results of image variation tasks, with corresponding quantitative results shown in \cref{table_imgvar}.}
    \label{fig_vertical_images}
\end{figure*}

Finally, along with the results on the CelebAHQ dataset shown in \cref{fig_comp_imgvar}, we provide additional visual results generated from the COCO validation dataset and ImageNet-R-TI2I to demonstrate that the proposed method consistently outperforms baseline models by faithfully reflecting the details of the given image prompt.
Starting with the COCO validation dataset (\cref{fig_imgvar_cocoval}), the proposed method accurately reproduces the details of the image prompt, such as the red dot pattern on a black tie, whereas the baselines simplify it to a plain red tie. Additionally, our method preserves intricate features like mural patterns, bedspread designs, and elongated light decorations in the generated images, while these elements are missing in images produced by the baselines. This trend is also observed in the results on the ImageNet-R-TI2I dataset (\cref{fig_imgvar_imgnetr}). The proposed method faithfully reflects textures, such as the pencil strokes in a sketch and the textures and colors of a painting. In contrast, IP-Adapter and SSR-Encoder exhibit color shifts, and RIVAL tends to smooth out the sketch textures or make paintings appear overly photorealistic.

\end{document}